\documentclass{article}
\usepackage{iclr2027_conference,times}
\usepackage[T1]{fontenc}
\usepackage{amsmath,amssymb,amsthm}
\usepackage{booktabs,array,tabularx}
\usepackage{algorithm,algpseudocode}
\usepackage{needspace}
\usepackage{placeins}
\usepackage{wrapfig}
\usepackage{graphicx,xcolor}
\usepackage{colortbl}
\usepackage{tikz}
\usetikzlibrary{arrows.meta,positioning,fit,calc,matrix,backgrounds}
\usepackage{microtype}
\usepackage{hyperref}
\usepackage{xurl}
\usepackage{graphicx}
\hypersetup{colorlinks=true,linkcolor=blue!55!black,citecolor=blue!55!black,urlcolor=blue!55!black}

\title{Is This Evidence Decision-Critical? \\Learning to Verify Rule-Governed Decisions}

\author{\hspace*{-\tabcolsep}%
\parbox{\dimexpr\textwidth-2\tabcolsep\relax}{\raggedright
Haoyang Zhang\textsuperscript{1,\textdagger}, Jianpeng Zhao\textsuperscript{1,\textdagger}, Qi Hao\textsuperscript{1}, Pengyang Wang\textsuperscript{1,*}\\[5pt]
\normalfont\small
\textsuperscript{1}State Key Laboratory of Internet of Things for Smart City and\\
\hphantom{\textsuperscript{1}}Institute of Smart City Technologies, University of Macau, Macau SAR, China
}
}

\DeclareRobustCommand{\stateSatisfied}{\ifmmode\text{Satisfied}\else Satisfied\fi}
\DeclareRobustCommand{\stateNotSatisfied}{\ifmmode\text{Not satisfied}\else Not satisfied\fi}
\DeclareRobustCommand{\stateUnknown}{\ifmmode\text{Unknown}\else Unknown\fi}
\DeclareRobustCommand{\decisionYes}{\ifmmode\text{Yes}\else Yes\fi}
\DeclareRobustCommand{\decisionNo}{\ifmmode\text{No}\else No\fi}
\DeclareRobustCommand{\decisionInsufficient}{\ifmmode\text{Insufficient evidence}\else\mbox{Insufficient} \mbox{evidence}\fi}

\definecolor{lfblue}{HTML}{275D8C}
\definecolor{lforange}{HTML}{B65D24}
\definecolor{lfink}{HTML}{20262D}
\definecolor{lfmuted}{HTML}{626B73}
\definecolor{lfline}{HTML}{D2D7DC}

\newcommand{\fignote}{\fontfamily{phv}\fontsize{7.6}{9.1}\selectfont}
\newcommand{\fighead}{\fontfamily{phv}\fontsize{9.3}{11}\selectfont\bfseries}
\DeclareMathSizes{8.4}{8}{6}{5}
\DeclareMathSizes{7.6}{8}{6}{5}
\tikzset{
  lf flow/.style={-{Latex[length=1.7mm,width=1.15mm]},line width=0.75pt,draw=lfmuted},
  lf block/.style={draw=lfline,line width=0.65pt,fill=white,align=center,
    inner sep=4pt,minimum height=0.85cm},
  lf label/.style={inner sep=0pt,align=center},
  lf title/.style={font=\fighead,anchor=west,inner sep=0pt,text=lfink},
  lf note/.style={font=\fignote,inner sep=0pt,text=lfmuted,align=center}
}
\newcommand{\method}{\textsc{InterPact}}
\newcommand{\framework}{\textsc{InterPact}}
\newcommand{\Cs}{\mathcal{S}}
\newcommand{\Ds}{\mathcal{D}}

\newcommand{\qa}{q^{\mathrm a}}
\newcommand{\db}{d^{\mathrm b}}
\newcommand{\da}{d^{\mathrm a}}
\newcommand{\cb}{c^{\mathrm b}}
\newcommand{\ca}{c^{\mathrm a}}
\newcommand{\ind}{\mathbf{1}}
\newcommand{\softmax}{\operatorname{softmax}}

\newcolumntype{Y}{>{\raggedright\arraybackslash}X}
\newcommand{\runresult}[2]{\mbox{#1\,{\fontsize{7}{9}\selectfont$\pm$\,#2}}}
\newlength{\resultmeanwidth}
\newlength{\resultsdwidth}
\newcommand{\alignresultcells}[2][00.00]{%
  \settowidth{\resultmeanwidth}{#1}%
  \settowidth{\resultsdwidth}{{\fontsize{7}{9}\selectfont #2}}%
  \renewcommand{\runresult}[2]{%
    \mbox{\makebox[\resultmeanwidth][r]{##1}\,%
      {\fontsize{7}{9}\selectfont$\pm$\,\makebox[\resultsdwidth][r]{##2}}}}%
}
\newcommand{\pointresult}[1]{\makebox[\resultmeanwidth][r]{#1}}

\newcommand{\bestresult}[1]{\textbf{#1}}
\newcommand{\secondresult}[1]{\underline{#1}}
\definecolor{tablegroupblue}{HTML}{EAF2F8}

\iclrfinalcopy
\hypersetup{pdftitle={Is This Evidence Decision-Critical? \\Learning to Verify Rule-Governed Decisions}, pdfauthor={Haoyang Zhang, Jianpeng Zhao, Qi Hao, Pengyang Wang}}

\begin{document}
\maketitle
\lhead{Preprint}
\begingroup
\renewcommand{\thefootnote}{\fnsymbol{footnote}}
\footnotetext[1]{Corresponding author.}
\footnotetext[2]{Equal contribution to this work.}
\endgroup

\begin{abstract}
Rule-based reasoning, as in eligibility checks and contract reviews, requires language models to assess evidence against individual conditions and combine their judgments under explicit rules.
Errors in evidence assessment can leave a decision unchanged, but misinterpreting or overlooking decision-critical evidence can reverse it.
Identifying such evidence allows more capable models to focus on checking the corresponding condition judgments, supporting accurate and safe decisions.
Recognizing the evidence's criticality requires understanding how evidence affects a condition judgment and how that judgment affects the decision.
% To achieve the goal, we introduce \framework{}, which internalizes these dependencies in the base model.
% To achieve the goal, we introduce \textbf{\framework}, which INTERvenes on evidence, measures its imPACT on the decision, and internalizes these dependencies in the base model.
To achieve the goal, we propose a \textbf{INTER}vention-based im\textbf{PACT} learning framework (\textbf{\framework}), which enables counterfactual verification of evidence criticality in rule-governed decisions.
Specifically, its evidence intervention constructor generates training pairs for a propagation verifier by editing case facts with a frozen language model while holding rules and non-target conditions fixed.
Human-reviewed labels record the resulting condition and decision changes, while complete state-to-decision mappings supervise consequences beyond the observed edit.
During training, the verifier weights learned conditional decision predictions by evidence-based condition probabilities through a fixed composition operation, propagating decision-change supervision into the base model.
At inference, the trained base model directly judges criticality from the original case and target evidence, without human or stronger-model supervision.
% On single-case evidence criticality verification across structured adaptations of ContractNLI and ShARC, \method{} achieves xx.xx\% accuracy, outperforming all x baselines.
On single-case evidence criticality verification over adapted rule-governed decision cases, \method{} achieves 68.28\% accuracy, outperforming all six baselines.
These results support learned decision sensitivity as a basis for prioritizing evidence checks.
% Our data and code are available at \url{URL}.

% TODO: Replace URL with the data and code release link.
\end{abstract}
\section{Introduction}
\label{sec:introduction}

\begingroup
\setlength{\intextsep}{3pt}
\setlength{\columnsep}{10pt}
\setlength{\abovecaptionskip}{3pt}
\begin{wrapfigure}{r}{0.56\textwidth}
\centering
\input{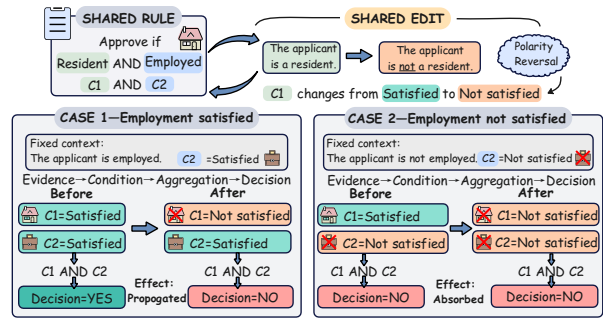}
\caption{\textbf{The same evidence change can have different decision consequences.}
A change in the target condition state changes the decision in one case
but leaves it unchanged in the other because the remaining condition states differ.}
\label{fig:motivation}
\end{wrapfigure}

\emph{Rule-based reasoning} 
involves assessing whether evidence supports, contradicts, or leaves individual conditions unresolved and combining these judgments under explicit rules to reach a decision in tasks such as eligibility checks \citep{saeidi2018sharc} and contract reviews \citep{koreeda2021contractnli}.
Errors in these judgments have different consequences: some leave the decision unchanged, whereas others can reverse it.
Identifying \emph{decision-critical evidence} allows more capable models to concentrate their checks on the judgments that affect the outcome, helping improve decision reliability and safety.

Criticality depends on the surrounding evidence and rule context.
Consider an application requiring both residency and employment, as illustrated in Figure~\ref{fig:motivation}.
A mistaken residency judgment can reverse approval when employment is satisfied, but leaves rejection unchanged when employment fails.
Even a local omission may leave a condition unchanged if other evidence still supports it.
A local evidence error therefore affects the outcome only when it changes a condition judgment that can change the decision.
Recognizing critical evidence requires understanding both dependencies.

A model can predict a case's decision correctly without recognizing which evidence changes could alter it.
Final-decision labels leave this sensitivity unspecified; adding condition labels as parallel targets still does not require the decision to depend on those judgments \citep{koh2020cbm}.
Learning criticality therefore calls for supervision of how evidence changes affect conditions and how those effects reach the decision.
Controlled interventions make these effects observable.
Holding rules and non-target conditions fixed, we compare the condition and decision before and after an evidence edit.
The resulting contrasts provide a signal for learning decision sensitivity.
This builds on counterfactual data augmentation \citep{kaushik2020counterfactual,qiu2024paircfr} and feedback \citep{huyuk2025feedback}, with supervision tracking both levels of consequence.
\par\WFclear
\endgroup

Our hypothesis is that learning these dependencies from interventions enables direct criticality judgments on original cases.
We introduce \framework{}, which trains a language model as a propagation verifier using supervision from an evidence intervention constructor.
The constructor edits case facts with a frozen language model and records condition and decision changes; human review assesses annotation quality.
Each edit yields one condition state.
Complete state-to-decision mappings therefore supply consequences for the other candidate states, teaching how alternative judgments would affect the decision.
During paired training, the verifier learns a decision distribution for each candidate condition state.
A fixed composition weights these predictions by evidence-based condition probabilities from a frozen estimator, making decision-change supervision depend on the intermediate judgments.
To test whether this learning supports direct verification, we query the trained language model alone with the original case and target evidence, asking whether any specified intervention could change the decision.
This requires no edited cases or human or stronger-model supervision.

On this single-case task across structured adaptations of ContractNLI and ShARC \citep{koreeda2021contractnli,saeidi2018sharc}, \method{} achieves 68.28\% accuracy and outperforms all six baselines.
% TODO: Fill in the single-case results together with the abstract; paired metrics are not substitutes.
% TODO: Compare training variants under the same direct-inference interface to test transfer; current paired ablations change both training and paired prediction.
Complementary paired ablations examine composition when predicting intervention outcomes.
With condition estimates and supervision held fixed, composition improves decision-change detection, supporting its role in connecting local judgments to their decision consequences.

Our contributions are threefold:
\begin{enumerate}
\renewcommand{\labelenumi}{(\theenumi)}
\item We introduce \framework{}, which turns the decision consequences of controlled evidence interventions into supervision for direct criticality judgments on original cases.
The learned verifier supports prioritizing evidence checks in rule-based reasoning.

\item We develop compositional training that links evidence assessment to rule-governed decisions, making decision-change supervision follow evidence--condition--decision dependencies.

\item Our evaluation separates direct criticality verification from paired analyses of intervention consequences.
Matched comparisons examine supervision and composition, showing that local judgment accuracy alone does not determine decision-change detection.
\end{enumerate}
% === End rewritten introduction by pywang ===

% CHECK: RW1
\section{Related Work}
\label{sec:related}

\paragraph{Evidence and Rule-based Reasoning.}
Studies of natural-language rule reasoning examine how conclusions follow
from stated facts and rules \citep{clark2020ruletaker,tafjord2021proofwriter},
including rule application in practical settings \citep{zhou2025rulearena}.
Evidence-linked inference examines how text supports judgments
\citep{koreeda2021contractnli}. Conversational rule reasoning and conditional
question answering examine how available evidence affects answers under
rules \citep{saeidi2018sharc,sun2022conditionalqa}.
Together, these studies motivate our setting: assessing evidence
against individual conditions and combining the resulting judgments under
explicit rules. \method{} builds on this setting by using controlled evidence
interventions to learn whether changing target evidence changes the target
condition state and whether that state change alters the rule-required decision.

% CHECK: RW2
\paragraph{Counterfactual Reasoning and Verification.}
Belief-R evaluates conclusion revision after additional premises
\citep{wilie2024belief}, while DeltaLogic studies responses to minimal
premise edits, including inertia and over-flips \citep{dhanda2026deltalogic}.
Counterfactual feedback uses consistency across paired questions for
fine-tuning \citep{huyuk2025feedback}, and PairCFR uses paired examples
for contrastive learning \citep{qiu2024paircfr}. Generative verification
and weighted logical inference assess candidate decisions
\citep{zhang2025genrm,kang2025r2guard}.
To verify evidence criticality, the evidence intervention constructor in
\method{} records the target condition state and rule-required decision
before and after each intervention. It also computes the decision for every
target condition state under the fixed rule context. These complete mappings
supervise consequences beyond the observed edits and allow conditional
predictions to be evaluated separately from final decisions.

% CHECK: RW3
\paragraph{Concept Bottlenecks and Intermediate Supervision.}
Concept bottleneck models distinguish prediction through intermediate
concepts from auxiliary concept supervision
\citep{koh2020cbm,desantis2026mechanistic}. Counterfactual extensions
support concept interventions \citep{dominici2025counterfactual}.
The \method{} Verifier learns context-dependent decision consequences
from complete rule-derived mappings and composes them with local condition
estimates. Hard Warrant constrains the known original condition-to-decision
relation. Matched ablations hold condition predictions fixed to test
composition beyond auxiliary supervision. Appendix~\ref{app:related}
extends these comparisons.

\section{Preliminaries}
\label{sec:task}

Rule-based reasoning determines a decision for a case under a governing rule.
The rule specifies conditions, each of which is assessed using the available
evidence. An aggregation function then combines the condition states to
answer the decision query. We represent the case as an instance
\begin{equation}
x = (r, Q, C, E, A),
\label{eq:reasoning-instance}
\end{equation}
where $r$ is the governing rule, $Q$ is the decision query,
$C=\{c_1,\ldots,c_n\}$ is the set of conditions, and $E$ is the available
evidence. The function $A$ explicitly represents the aggregation logic
of $r$ as a mapping from condition states to a final decision.
For each condition $c_i$,
$E_i \subseteq E$ denotes the set of evidence units associated with that
condition. If $E_i$ contains $m_i$ evidence units, then
$E_i=\{e_{i1},\ldots,e_{im_i}\}$, where $e_{ij}$ denotes an
evidence unit associated with $c_i$.

Given an instance $x$, each condition $c_i$ has a state
$c_i(x)$ in the three-way state space
$\mathcal{S}=\{\textsc{Satisfied},\textsc{Not Satisfied},\textsc{Unknown}\}$.
\textsc{Satisfied} means that the evidence supports condition $c_i$.
\textsc{Not Satisfied} means that the evidence supports the negation of $c_i$.
\textsc{Unknown} means that the evidence is insufficient to determine
whether $c_i$ holds.
The condition-state vector is
\begin{equation}
C(x) = \bigl(c_1(x),\ldots,c_n(x)\bigr).
\label{eq:condition-assignment}
\end{equation}
The aggregation function maps this vector to the decision required by $r$:
\begin{equation}
D(x) = A\bigl(C(x)\bigr),
\label{eq:reference-decision}
\end{equation}
where $D(x)\in\mathcal{D}
=\{\textsc{Yes},\textsc{No},\textsc{Insufficient Evidence}\}$.
\textsc{Yes} and \textsc{No} denote positive and negative answers
to $Q$ under $r$.
\textsc{Insufficient Evidence} means that the available evidence
does not determine an answer.

% xxx per point,shorten as qq (xx\%->xx\%)

\section{\method{}}
\label{sec:method}

\paragraph{Overview.}
To identify decision-critical evidence from a single structured case,
\framework{} learns from controlled evidence interventions. The framework
has two components. (1) The \textbf{Evidence Intervention Constructor}
applies allowed interventions to a target evidence unit. During each
intervention, the constructor keeps the governing rule and other evidence
fixed. If an allowed intervention can change the decision, the target
evidence unit is decision-critical. The constructor produces labeled
intervention pairs and complete condition-to-decision mappings. However,
a change in the target condition state does not always change the
decision. (2) The \textbf{\method{} Verifier} therefore
estimates the target condition state and predicts the decision for each
possible state. The intervention pairs and mappings supervise verifier
training. At inference, the trained language model
judges whether the target evidence is decision-critical in the original
case.

% CHECK: U73
\begin{figure}[t]
\centering
\resizebox{0.9\textwidth}{!}{%
\input{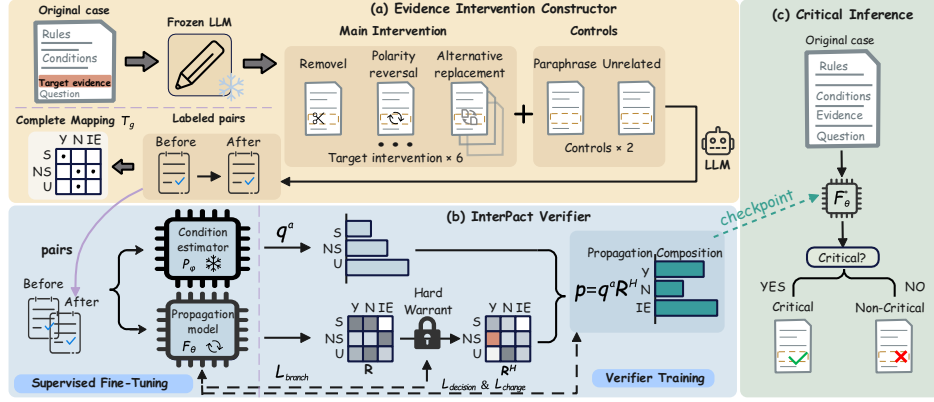}}
\caption{\textbf{Overview of \framework{}.}
(a) The Evidence Intervention Constructor produces labeled intervention
pairs and complete condition-to-decision mappings.
(b) The \method{} Verifier learns from these outputs through propagation
composition and Hard Warrant.
(c) The trained language model directly judges evidence criticality from
the original case and target evidence unit.}
\label{fig:model}
\end{figure}

% CHECK: END-U73

\subsection{Evidence Intervention Constructor}
\label{sec:construction}

The constructor takes an original case $x_g$ and a target evidence unit
$e_g\in E_t$ associated with target condition $c_t$ as input.
The case, target evidence unit, and target condition together define a
\emph{root} $g$. The constructor uses the annotated condition states
$C(x_g)$ and decision $D(x_g)$ during offline construction. For each root,
the constructor generates controlled evidence interventions, determines
the resulting target condition states and decisions, and computes a
complete condition-to-decision mapping.

\subsubsection{Controlled Evidence Interventions}
To examine how changes to a target evidence unit affect the
target condition state and final decision, the constructor uses six
target operations: removal, polarity reversal, alternative replacement,
irrelevant replacement, weakening, and strengthening.
While these operations test possible condition and decision changes,
paraphrase and distractor insertion serve as invariance controls.
For removal, the constructor deletes the target evidence unit directly.
For the other operations, it prompts a frozen \mbox{GPT-5.4} model to
rewrite the target unit or insert a distractor as required by the operation.
The constructor requires every intervention to preserve the governing
rule, decision query, condition descriptions, aggregation function, and
existing non-target evidence. If it can't produce a suitable intervention,
the constructor abstains. Appendix~\ref{app:interventions} defines the
operations and reports construction checks and retained pair counts.

\subsubsection{Intervention Outcome Determination and Records}
The constructor excludes interventions that fail to follow the
specified operation or change non-target condition states. Because
other evidence may preserve the target condition state, it uses model
assistance to judge the state of $c_t$ from the full evidence after
each accepted intervention. The model does not see the expected label
or operation name. The constructor forms the condition-state vector
after the intervention by replacing only the target entry of $C(x_g)$
with the judged state. It then uses $A$ to compute the decision from the condition states,
avoiding a separate model prediction that could conflict with the rule.

For accepted intervention $i$, $t_i$ indexes the target condition
and $j_i$ indexes the target evidence unit $e_{t_i j_i}$ within that
condition. The constructor records
$z_i=(x_i^{\mathrm b},x_i^{\mathrm a},t_i,j_i)$,
where $x_i^{\mathrm b}$ and $x_i^{\mathrm a}$ are the cases before
and after the intervention. It also records the target condition state
and decision before and after the intervention, and whether the decision
changes. An operation may yield no accepted intervention for a given
original case and target evidence unit.
Sampling across many original cases and target evidence units still
provides varied condition and decision labels for training the
\method{} Verifier.

\subsubsection{Complete Condition-to-Decision Mappings}
For each accepted intervention, the constructor judges one target
condition state. To obtain decisions for all three states, the
constructor computes
\begin{equation}
T_g(c)=A_g\!\left(C(x_g)_{t\leftarrow c}\right),
\qquad c\in\mathcal{S}.
\label{eq:target}
\end{equation}
Here $A_g$ is the aggregation function of the original case $x_g$.
The condition-state vector $C(x_g)_{t\leftarrow c}$ sets the state of
target condition $c_t$ to $c$ and keeps all other states from $C(x_g)$.
This calculation does not edit the evidence.

For pair $i$, $g(i)$ identifies the original case and target evidence unit. Thus $x_i^{\mathrm b}=x_{g(i)}$,
and pairs with the same $g(i)$ share $T_{g(i)}$. The mapping serves as
a training target and evaluation reference for conditional decision
predictions, including states absent from sampled interventions.
The verifier does not receive $T_g$ as input.
Appendix~\ref{app:decision-labels} reports how the complete mappings
extend supervision beyond the sampled interventions.

\subsection{\method{} Verifier}
\label{sec:verifier}

A target condition change need not change the decision,
because its effect depends on the rule and other conditions.
The \method{} Verifier therefore separates condition estimation from
conditional decision prediction. Figure~\ref{fig:model}(b) shows the
verifier's computation during paired training.

For pair $z_i$, write $t=t_i$ for the target condition index.
Let $\cb_i=c_t(x_i^{\mathrm b})$ and $\db_i=D(x_i^{\mathrm b})$ denote
the target state and decision before the intervention. These annotations
are supplied during paired training and paired evaluation. They are
excluded from direct criticality inference, described in
Section~\ref{sec:direct-criticality}.

\subsubsection{Condition estimation and conditional decision prediction}
\label{sec:group}

Using the case after the intervention, a condition estimator $P_\phi$
assigns a probability to each state of the target condition:
$\qa_i[c]=P_\phi(c\mid x_i^{\mathrm a},c_t)$, for $c\in\Cs$.
The row vector $\qa_i$ follows the state order in Section~\ref{sec:task}
and sums to one. It retains uncertainty about the target condition.

The aggregation function is known, but the non-target condition states
are not given to the verifier. For each possible target condition state
$c$, a language model $F_\theta$ uses the pair record $z_i$ and $c$ to
predict a decision distribution. The governing rule and non-target
evidence remain fixed across these predictions.
The model produces logits $Z_i\in\mathbb R^{3\times3}$ and normalizes each
row to obtain $R_i=\softmax_{\mathrm{row}}(Z_i)$.
Rows index target states in $\Cs$ and columns index decisions in $\Ds$.
The entry $R_i[c,d]$ gives the predicted probability of decision $d$
for target condition state $c$.

\subsubsection{Propagation composition with Hard Warrant}
\label{sec:drift}
\label{sec:hardwarrant}

Under the fixed governing rule and non-target evidence, the original
target state $\cb_i$ must yield the original decision $\db_i$.
The model may otherwise predict a different decision for this known state.
\emph{Hard Warrant} fixes row $R_i[\cb_i,:]$ to $\mathbf e_{\db_i}$,
the one-hot decision vector for $\db_i$:
\begin{equation}
 R_i^H[c,:]=
 \begin{cases}
 \mathbf e_{\db_i}, & c=\cb_i,\\
 R_i[c,:], & c\ne\cb_i.
 \end{cases}
 \label{eq:hard-warrant}
\end{equation}
This constraint uses only the annotated target state and decision before
the intervention. It does not require the complete mapping $T_{g(i)}$.

Propagation composition weights the three rows of $R_i^H$ by $\qa_i$
to obtain the decision distribution $p_i$ after the intervention:
\begin{equation}
 p_i[d]=\sum_{c\in\Cs}\qa_i[c]R_i^H[c,d],
 \qquad p_i=\qa_i R_i^H.
 \label{eq:projection}
\end{equation}

For pair $i$, the predicted probability of a decision change is
\begin{equation}
 s_i=1-p_i[\db_i].
 \label{eq:criticality-probability}
\end{equation}
Under Hard Warrant, the row for $\cb_i$ cannot contribute to a decision
change. A change in the target condition may still leave the decision
unchanged, so $s_i$ can remain low.
Hard Warrant preserves the known relation but cannot correct errors in
condition estimation or decision predictions for other target states.
Appendix~\ref{app:proof} derives these limits.

\subsubsection{Training objective}

To learn target condition states and how they affect decisions, training
proceeds in two stages. Supervised fine-tuning first trains a model to
predict the target condition state and decision from a case after an intervention.
The resulting parameters initialize the condition estimator $P_\phi$ and
language model $F_\theta$. The second stage updates $\theta$ while
keeping $\phi$ fixed.

Let $\ca_i=c_t(x_i^{\mathrm a})$ and $\da_i=D(x_i^{\mathrm a})$ denote
the target state and decision after the intervention.
Because accepted interventions leave non-target condition states unchanged,
$\db_i=T_{g(i)}(\cb_i)$ and $\da_i=T_{g(i)}(\ca_i)$.
The decision-change label for pair $i$ is $y_i=\ind[\da_i\ne\db_i]$.
Branch cross-entropy $\mathcal L_{\mathrm{branch}}$ supervises every row
of $R_i$ before Hard Warrant with the decision $T_{g(i)}(c)$.
Decision cross-entropy $\mathcal L_{\mathrm{decision}}$ supervises $p_i$
with $\da_i$, while binary cross-entropy $\mathcal L_{\mathrm{change}}$
supervises $s_i$ with $y_i$. Both use outputs computed after Hard Warrant.
The training objective is
\begin{equation}
 \mathcal L_{\mathrm{train}}
 =\mathcal L_{\mathrm{decision}}
  +\lambda_{\mathrm{branch}}\mathcal L_{\mathrm{branch}}
  +\lambda_{\mathrm{change}}\mathcal L_{\mathrm{change}}.
 \label{eq:objective-summary}
\end{equation}
The coefficients $\lambda_{\mathrm{branch}}$ and
$\lambda_{\mathrm{change}}$ weight the branch and change losses.
Propagation composition and Hard Warrant have no trainable parameters.
The branch loss updates $F_\theta$ directly, while gradients from the
decision and change losses pass through these fixed computations to
update $F_\theta$.
Appendix~\ref{app:objective} gives the loss definitions, coefficient values,
and paired prediction procedure. Algorithm~\ref{alg:training} describes training.

\subsubsection{Direct evidence criticality inference}
\label{sec:direct-criticality}

The paired predictions above concern supplied intervention pairs.
Direct evidence criticality inference asks whether an allowed change to
the target evidence unit could alter the decision for the original case.
For root $g$, let $\mathcal O_g$ contain the valid interventions to $e_g$
that follow the six target operations in Section~\ref{sec:construction}.
Each intervention keeps the governing rule, decision query, condition
descriptions, aggregation function, and non-target evidence fixed. It also
leaves non-target condition states unchanged.
Allowed interventions are specified independently of their outcomes.
For each $o\in\mathcal O_g$, let $x_{g,o}^{\mathrm a}$ denote the case
after the intervention.
The reference decisions before and after the intervention are
$\db_g=D(x_g)$ and $\da_{g,o}=D(x_{g,o}^{\mathrm a})$.
The reference label is
\begin{equation}
 y_g=\ind\!\left[\exists o\in\mathcal O_g:
                  \da_{g,o}\ne\db_g\right].
 \label{eq:root-criticality}
\end{equation}
For evaluation, a decision-changing accepted intervention establishes a
positive label. Without an observed decision change, a negative label
requires either a complete mapping that rules out a change or review of
the rule and case evidence. Missing interventions alone do not establish
a negative label. These negative labels do not require an exhaustive
search of $\mathcal O_g$. The reference labels $y_g$ evaluate direct inference,
while the pair labels $y_i$ supervise training.

Let $\theta^*$ denote the checkpoint selected from the second stage on
development pairs (Appendix~\ref{app:selection}). At inference, a
natural-language prompt gives $F_{\theta^*}$ the original case $x_g$
and target evidence unit $e_g$. It asks whether changing $e_g$ could
change the decision under the governing rule.
The answers \texttt{Yes} and \texttt{No} concern evidence criticality,
not the case decision $D(x_g)$. They give $\hat y_g=1$ and $\hat y_g=0$,
respectively.
Other responses are recorded as invalid outputs.
The prompt contains no condition-state labels, decision labels,
edited case, or reference mapping.
Direct inference runs only $F_{\theta^*}$ and requires no intervention
construction or human review. Evaluation compares $\hat y_g$ with $y_g$.
The score $s_i$ applies only to paired prediction.

\section{Experiments}
\label{sec:experiments}

This section examines four questions: direct criticality judgments from a
single case, the training components that improve these judgments, decision
prediction for each target condition state, and decision-change prediction
for intervention pairs.

\subsection{Experimental setup}
\label{sec:experimental-setup}

\paragraph{Data and Reference Labels.}
We use 6,913 accepted intervention pairs from structured adaptations of
ContractNLI and ShARC \citep{koreeda2021contractnli,saeidi2018sharc}.
The cases use three-valued conjunction or disjunction, and related documents,
rule families, and near-duplicates remain in the same split.
Direct evaluation covers 207 test roots: 85 critical and 122 non-critical.
The criticality labels concern the six target operations in
Section~\ref{sec:construction}.
An accepted intervention that changes the decision establishes a positive label.
A negative label is assigned when the complete mapping gives the original
decision for every target state, or when reviewers judge from the rule and
case evidence that no allowed edit changes the decision.
Missing interventions alone do not justify a negative label.
Appendix~\ref{app:interventions} details construction, operation counts,
and data splits; Appendix~\ref{app:criticality-labels} explains the
reference-label review.

\paragraph{Baselines and Evaluation.}
The baselines include Qwen3.5-4B without task-specific training, After-State SFT, and Predicted State Execution (PSE). We also adapt PairCFR~\citep{qiu2024paircfr}, GenRM~\citep{zhang2025genrm}, and R$^2$-Guard~\citep{kang2025r2guard} as reasoning baselines. All trainable models use Qwen3.5-4B with LoRA~\citep{hu2022lora}.
For direct evaluation, every model receives the same prompt containing the
original case with its target evidence unit identified.
PSE and R$^2$-Guard use only their trained language models; paired
diagnostics use their complete systems.
Trainable checkpoints are selected on paired development data without
training on root criticality labels. We report Accuracy, Balanced Accuracy
(BA), and Macro-F1 over test roots, counting invalid answers as errors.
Unless stated otherwise, results are mean $\pm$ sample standard deviation
over three runs. Appendix~\ref{app:baselines} describes the baseline
adaptations, and Appendix~\ref{app:implementation} gives training and
reporting details.

\subsection{RQ1: Can the model identify critical evidence from a single case?}
\label{sec:main-results}

This section evaluates direct criticality judgments from the original case with
the target evidence unit identified. At inference, the models receive no
edited case or reference labels. Compared with the strongest baseline for
each metric, \method{} improves mean accuracy by 4.51 percentage points (pp)
over GenRM and balanced accuracy by 5.76 pp over PairCFR
(Table~\ref{tab:direct-criticality}). These gains show that the trained
language model improves direct criticality judgments from a single case, the
inference setting targeted by the framework.

% Values are sourced from tables/direct_criticality_results.json.
% Missing results and standard deviations are not imputed.
\begin{table}[htbp]
\caption{Direct criticality on 207 roots (\%).
\textbf{Bold}/\underline{Underline}: best/second-best classification scores.}
\label{tab:direct-criticality}
\centering\small
\alignresultcells{0.00}
\newcommand{\directmean}[1]{\pointresult{#1}\hphantom{\,{\fontsize{7}{9}\selectfont$\pm$\,0.00}}}
\setlength{\tabcolsep}{4pt}
\begin{tabular}{lcccc}
\toprule
 & \multicolumn{3}{c}{Criticality Judgment $\uparrow$} & \\
\cmidrule(lr){2-4}
Method & Acc. & BA & Macro-F1 & Invalid rate $\downarrow$\\
\midrule
\rowcolor{tablegroupblue}[0pt][0pt]
\multicolumn{5}{l}{\emph{LLM-based baselines}}\\
Qwen3.5-4B & \runresult{60.07}{1.55} & \runresult{58.69}{2.15} & \runresult{58.64}{2.08} & \directmean{0.00}\\
After-State SFT & \runresult{62.64}{2.23} & \runresult{57.13}{3.70} & \runresult{54.40}{6.46} & \directmean{0.00}\\
Predicted State Execution & \runresult{51.05}{1.95} & \runresult{51.63}{1.76} & \runresult{51.30}{2.26} & \runresult{1.61}{1.39}\\
\midrule
\rowcolor{tablegroupblue}[0pt][0pt]
\multicolumn{5}{l}{\emph{Adapted reasoning baselines}}\\
PairCFR & \runresult{59.74}{2.28} & \runresult{\secondresult{60.85}}{1.58} & \runresult{\secondresult{59.60}}{2.08} & \directmean{0.00}\\
% GenRM/PSE/R2-Guard metrics recomputed from two-run summaries plus supplied third runs.
% Pooled and source results now consistently aggregate three runs with sample SD.
GenRM & \runresult{\secondresult{63.77}}{0.84} & \runresult{57.19}{0.92} & \runresult{53.48}{1.23} & \directmean{0.00}\\
R$^2$-Guard & \runresult{52.98}{1.12} & \runresult{53.33}{0.64} & \runresult{52.95}{1.17} & \runresult{0.97}{0.84}\\
\midrule
\method{} & \runresult{\bestresult{68.28}}{2.01} & \runresult{\bestresult{66.61}}{1.96} & \runresult{\bestresult{66.78}}{2.01} & \directmean{0.00}\\
\bottomrule
\end{tabular}
\end{table}                                                                            

After-State SFT improves mean accuracy over Qwen3.5-4B by 2.57 pp
but reduces balanced accuracy by 1.56 pp
(Table~\ref{tab:direct-criticality}). Higher accuracy alone therefore
does not ensure more balanced criticality judgments. Across sources,
\method{} improves mean balanced accuracy by 3.90 pp over PairCFR on
ContractNLI and by 1.66 pp over GenRM on ShARC
(Figure~\ref{fig:direct-criticality-by-source} and
Table~\ref{tab:direct-criticality-by-source} in
Appendix~\ref{app:direct-by-source}). Both sources show a mean gain,
but the gains differ in size and the results vary across runs. Since
direct inference uses only the trained language model, we examine which
training components contribute to these gains.

\subsection{RQ2: Which training components improve direct criticality judgments?}
\label{sec:controls}

We compare \method{} with four matched ablations
(Appendix~\ref{app:controls}). All variants answer from a single case.
Propagation composition and Hard Warrant are used in paired training and
checkpoint selection, but not at direct inference.
\method{} has higher mean balanced accuracy and Macro-F1 than every ablation
(Table~\ref{tab:direct-ablations}). In the composition ablation, branch loss
still supervises the conditional decision predictions, but these predictions
do not determine the decision. Removing propagation composition lowers
balanced accuracy by 9.76 pp, supporting its contribution beyond branch
supervision.

% Values are sourced from tables/direct_criticality_results.json.
% Missing results and standard deviations are not imputed.
\begin{table}[htbp]
\caption{Training ablations on direct criticality (\%). \textbf{Bold} marks the best mean.}
\label{tab:direct-ablations}
\centering\small
\alignresultcells{0.00}
\setlength{\tabcolsep}{5pt}
\begin{tabular}{lccc}
\toprule
Method & Acc. $\uparrow$ & BA $\uparrow$ & Macro-F1 $\uparrow$\\
\midrule
w/o propagation composition & \runresult{62.32}{0.84} & \runresult{56.85}{1.07} & \runresult{54.80}{1.52}\\
w/o branch loss & \runresult{63.77}{0.48} & \runresult{60.46}{0.60} & \runresult{60.36}{0.66}\\
w/o Hard Warrant & \runresult{58.78}{0.74} & \runresult{51.65}{0.71} & \runresult{45.66}{0.93}\\
w/o change loss & \runresult{58.94}{0.49} & \runresult{54.16}{0.72} & \runresult{52.65}{1.11}\\
\midrule
\method{} & \runresult{\bestresult{68.28}}{2.01} & \runresult{\bestresult{66.61}}{1.96} & \runresult{\bestresult{66.78}}{2.01}\\
\bottomrule
\end{tabular}
\end{table}

Removing branch loss lowers balanced accuracy by 6.15 pp, indicating
that complete condition-to-decision mappings provide useful supervision.
Removing Hard Warrant or change loss lowers it by 14.96 and 12.45 pp,
respectively. These decreases support preserving the known relation between
the original target state and decision and explicitly supervising decision
changes in paired training. The effect of branch loss differs by source:
removing it raises balanced accuracy by 1.00 pp on ContractNLI but lowers
it by 13.62 pp on ShARC
(Table~\ref{tab:direct-criticality-by-source}). Branch loss appears useful
on ShARC.

\Needspace{18\baselineskip}
\subsection{RQ3: Can the verifier predict the decision for each target condition state?}
\label{sec:conditional-results}

\begingroup
\setlength{\intextsep}{5pt}
\setlength{\columnsep}{9pt}
\setlength{\abovecaptionskip}{5pt}
\begin{wrapfigure}{r}{0.44\textwidth}
\centering
\includegraphics[width=0.90\linewidth]{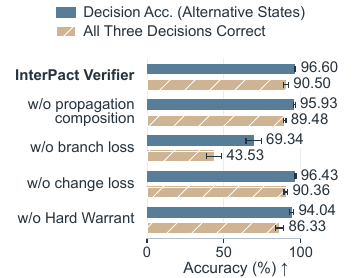}
\caption{Conditional decision accuracy before Hard Warrant.
Scores are averaged within roots and then equally across roots.}
\label{fig:branch-accuracy}
\end{wrapfigure}

On 978 target pairs, we evaluate conditional decision predictions to test
supervision from complete condition-to-decision mappings. The
\mbox{\method{}} Verifier
receives both evidence versions and the target condition state and decision
before the intervention, but no after-intervention labels or reference mapping.
We compare each decision predicted by $R_i$ with the corresponding
decision in $T_{g(i)}$ before applying Hard Warrant.
\emph{Decision Acc.\ (Alternative States)} covers the two states other
than the original target state, while \emph{All Three Decisions Correct}
requires correct predictions for all three states
(Appendix~\ref{app:metric-details}).

Compared with the variant without branch loss, the \mbox{\method{}} Verifier
gains 27.26 pp in alternative-state decision accuracy and 46.97 pp in
All Three Decisions Correct (Figure~\ref{fig:branch-accuracy}). Both
measures evaluate $R_i$ before Hard Warrant, supporting the role of complete
mapping supervision. Removing propagation composition lowers All Three Decisions
Correct by 1.02 pp but lowers direct balanced accuracy by 9.76 pp
(Figure~\ref{fig:branch-accuracy}; Table~\ref{tab:direct-ablations}).
Thus, accurate conditional decision predictions alone do not explain the
gain in direct judgments. Using them in the paired decision computation
also matters.
\WFclear
\endgroup

\subsection{RQ4: Can the paired verifier correctly predict decision changes?}
\label{sec:paired-results}

On the same 978 target pairs, we evaluate the paired verifier's decision
after an intervention and its decision-change prediction. Decision Change Balanced Acc.\
measures change detection; decision
and condition accuracy assess the labels after the intervention where
available. Each model's decision-change threshold is selected on
development pairs and fixed for testing (Appendix~\ref{app:selection}).
Relative to GenRM, the strongest baseline on both decision measures, the
\mbox{\method{}} Verifier improves decision-change balanced accuracy by
3.18 pp and decision accuracy by 2.96 pp (Table~\ref{tab:main-results}).
The paired verifier thus improves change detection and decision accuracy.

% CHECK: U55
% Baseline mean/SD from evidence/tutor_baseline_statistics.json. No pooled predictions.
\begin{table}[htbp]
\caption{Paired predictions on 978 target interventions (\%).
Bold/Underline: best/second-best means; N/A: unavailable output.}
\label{tab:main-results}
\centering\small
\alignresultcells{0.00}
\setlength{\tabcolsep}{4pt}
\begin{tabular}{lccccc}
\toprule
 & Decision Change $\uparrow$
 & \multicolumn{2}{c}{Decision $\uparrow$}
 & \multicolumn{2}{c}{Condition $\uparrow$}\\
\cmidrule(lr){2-2}\cmidrule(lr){3-4}\cmidrule(lr){5-6}
Method & Balanced Acc. & Acc. & Macro-F1 & Acc. & Macro-F1\\
\midrule
\rowcolor{tablegroupblue}[0pt][0pt]
\multicolumn{6}{l}{\emph{LLM-based baselines}}\\
Prompted Qwen3.5-4B & \runresult{82.59}{1.12} & \runresult{73.31}{0.37} & \runresult{72.00}{0.41} & \runresult{64.31}{0.64} & \runresult{64.72}{0.64}\\
After-State SFT & \runresult{87.22}{0.51} & \runresult{86.95}{0.33} & \runresult{85.96}{0.28} & \runresult{83.27}{0.33} & \runresult{83.24}{0.38}\\
Predicted State Execution & \runresult{86.22}{1.23} & \runresult{85.55}{1.25} & \runresult{84.28}{1.39} & \runresult{83.47}{0.79} & \runresult{83.43}{0.70}\\
\midrule
\rowcolor{tablegroupblue}[0pt][0pt]
\multicolumn{6}{l}{\emph{Adapted reasoning baselines}}\\
% PairCFR SDs updated from author-verified values on 2026-09-26.
PairCFR & \runresult{83.75}{0.62} & \runresult{82.11}{0.47} & \runresult{81.43}{0.41} & N/A & N/A\\
GenRM & \runresult{\secondresult{87.65}}{0.93} & \runresult{\secondresult{88.45}}{0.72} & \runresult{\secondresult{87.53}}{0.90} & N/A & N/A\\
R$^2$-Guard & \runresult{87.25}{1.06} & \runresult{87.56}{0.77} & \runresult{86.55}{0.85} & \runresult{\secondresult{84.05}}{2.24} & \runresult{\secondresult{83.94}}{2.34}\\
\midrule
\method{} & \runresult{\bestresult{90.83}}{0.73} & \runresult{\bestresult{91.41}}{0.67} & \runresult{\bestresult{90.58}}{0.71} & \runresult{\bestresult{84.39}}{0.26} & \runresult{\bestresult{84.39}}{0.32}\\
\bottomrule
\end{tabular}
\end{table}

% CHECK: END-U55

Targets are \emph{invariant} when the condition state stays fixed,
\emph{absorbed} when it changes without changing the decision, and
\emph{propagated} when both the condition state and decision change. The matched
ablations share condition estimates within each run, allowing the
comparison to focus on decision prediction. Relative to the variant
without propagation composition, the full verifier improves
decision-change balanced accuracy by 2.31 pp and decision accuracy on
propagated targets by 5.28 pp (Table~\ref{tab:controls}). Composition
therefore helps when a condition change must alter the decision.
Compared with the variant without Hard Warrant, the full verifier gains
4.17 pp in decision accuracy on absorbed targets and 4.16 pp overall,
but loses 0.74 pp on propagated targets (Table~\ref{tab:controls}).
Hard Warrant appears to preserve unchanged decisions but may make required
changes harder to predict; Appendix~\ref{app:errors} examines this at a
fixed checkpoint.

% CHECK: U63
% Repeated ablations use their dedicated run-statistics files.
\begin{table}[htbp]
\caption{Matched ablations on 978 target pairs (\%).
Variants share condition estimates. The first three metrics use all pairs;
the last two report accuracy on absorbed and propagated pairs, respectively.}
\label{tab:controls}
\centering\small
\alignresultcells{0.00}
\setlength{\tabcolsep}{2.5pt}
\begin{tabular}{lccccc}
\toprule
 & Decision Change $\uparrow$
 & \multicolumn{4}{c}{Decision $\uparrow$}\\
\cmidrule(lr){2-2}\cmidrule(lr){3-6}
Method & Balanced Acc. & Acc. & Macro-F1 & Absorbed Acc.
 & Propagated Acc.\\
\midrule
w/o propagation composition & \runresult{88.52}{1.22} & \runresult{90.25}{0.26} & \runresult{89.47}{0.16} & \runresult{\secondresult{92.32}}{1.58} & \runresult{76.51}{3.53}\\
w/o branch loss & \runresult{\secondresult{89.19}}{1.19} & \runresult{90.63}{0.82} & \runresult{89.75}{0.71} & \runresult{91.28}{1.13} & \runresult{79.30}{2.02}\\
w/o change loss & \runresult{88.88}{0.63} & \runresult{\secondresult{90.83}}{0.87} & \runresult{\secondresult{89.99}}{0.80} & \runresult{91.54}{2.00} & \runresult{80.76}{1.83}\\
w/o Hard Warrant & \runresult{87.70}{0.94} & \runresult{87.25}{0.77} & \runresult{86.32}{0.83} & \runresult{89.32}{2.77} & \runresult{\bestresult{82.53}}{0.25}\\
\method{} & \runresult{\bestresult{90.83}}{0.73} & \runresult{\bestresult{91.41}}{0.67} & \runresult{\bestresult{90.58}}{0.71} & \runresult{\bestresult{93.49}}{1.37} & \runresult{\secondresult{81.79}}{2.26}\\
\bottomrule
\end{tabular}
\end{table}

% CHECK: END-U63

Relative to GenRM,
decision accuracy improves by 5.60 pp on ContractNLI but only 0.14 pp on
ShARC, where GenRM has 0.10 pp higher decision-change balanced accuracy
(Table~\ref{tab:source-controls} in
Appendix~\ref{app:paired-source-controls}). On 381 paraphrase and
distractor insertion controls, the verifier incorrectly predicts decision
changes 6.30 pp more often than GenRM
(Table~\ref{tab:source-controls}). These results support better
decision-change predictions on target interventions, while false changes
on controls remain a limitation.

\FloatBarrier

% CHECK: C1
\Needspace{14\baselineskip}
\section{Conclusion}

% \framework{} trains a language model to judge evidence criticality directly
% from a single structured case. Controlled intervention pairs and complete
% condition-to-decision mappings provide supervision. Propagation composition
% and Hard Warrant connect condition estimates to decision predictions during
% training. At inference, the trained model judges the target evidence without
% supplied condition or decision labels or online intervention construction.
% Experiments show improved direct judgments, and matched ablations support
% the benefit of composition beyond auxiliary supervision.
% Training and paired evaluation assume known original target states and
% decisions. Evaluation is limited to two sources, three-valued AND/OR rules,
% and single-condition interventions. Future work will examine whether these
% judgments improve evidence review and decision reassessment.

Identifying decision-critical evidence in rule-based reasoning requires tracing
how evidence affects condition states and how these states affect decisions.
Final decision labels alone do not reveal these dependencies. The Evidence
Intervention Constructor produces 6,913 accepted pairs and complete
condition-to-decision mappings for training. During paired training, the
\method{} Verifier learns from this supervision through propagation composition
and Hard Warrant. At inference, the trained model judges criticality from the
original case and target evidence, achieving 68.28\% accuracy, 4.51 percentage
points above GenRM. These results support intervention-derived process
supervision for direct criticality judgments.

\label{page:main-end}

\section{Limitations and Future Work}

Our experiments study evidence criticality in a controlled setting with textual
evidence and three-valued conjunction or disjunction. Offline construction uses
annotated condition states and decisions for original cases. Future work will
explore other aggregation rules and interventions involving multiple evidence
units, reduce annotation requirements for construction, and examine how direct
criticality judgments support evidence review.

\section*{AI use statement}

% CHECK: U52
We used generative AI tools to assist data construction and annotation,
language editing, and layout preparation. The authors take responsibility
for the final content of this work, including all text, claims, and artifacts
produced with AI assistance.

% CHECK: END-U52

\bibliography{references}
\bibliographystyle{iclr2027_conference}

\clearpage
\raggedbottom
\appendix
\section{Properties of Hard Warrant}
\label{app:proof}

These properties apply to paired training and evaluation. Direct
criticality inference in Section~\ref{sec:direct-criticality} does not
execute Hard Warrant.

For root $g$, the original target state $\cb_g=c_t(x_g)$ and decision
$\db_g=D(x_g)$ satisfy
\begin{equation}
 T_g(\cb_g)=\db_g,\qquad R_g^*[\cb_g,:]=\mathbf e_{\db_g},
 \label{eq:warrant}
\end{equation}
Here $\mathbf e_v$ is the one-hot vector for a state or decision $v$.
The matrix $R_g^*$ is the one-hot table for the reference mapping $T_g$.
Hard Warrant fixes this row of the learned table. For
$p_i^{\mathrm{raw}}=\qa_i R_i$ and $p_i^H=\qa_i R_i^H$,
\begin{equation}
 p_i^{\mathrm{raw}}-p_i^H
 =\qa_i[\cb_i]\bigl(R_i[\cb_i,:]-\mathbf e_{\db_i}\bigr).
 \label{eq:drift}
\end{equation}
Thus an error in this row contributes in proportion to the predicted
probability that the target condition remains unchanged.

Hard Warrant replaces only $R[\cb,:]$ with $\mathbf e_{\db}$ in the
row-stochastic table $R$, where $\cb\in\Cs$ and $\db\in\Ds$ are the
supplied original state and decision. The resulting $R^H$ remains
row-stochastic, and $p^H=\qa R^H$ is normalized for any condition
probability row vector $\qa$.
If $\qa=\mathbf e_{\cb}$, then $p^H=\mathbf e_{\db}$; if
$\qa=\mathbf e_c$ for $c\ne\cb$, then $p^H=R[c,:]$.

The row fixed by Hard Warrant contributes no probability to a decision
change. Expanding Equation~\ref{eq:criticality-probability} therefore gives
\begin{equation}
 s_i=\sum_{c\ne\cb_i}\qa_i[c]\bigl(1-R_i[c,\db_i]\bigr).
 \label{eq:hardscore}
\end{equation}
Each term combines the probability of an alternative target state with
the probability that this state leads to a different decision.
Since $0\leq 1-R_i[c,\db_i]\leq1$,
$s_i\leq1-\qa_i[\cb_i]$. Confidence in a condition change alone therefore
does not determine the probability of a decision change.

For $s^{\mathrm{raw}}=1-p^{\mathrm{raw}}[\db]$ and
$s^H=1-p^H[\db]$, Equation~\ref{eq:drift} gives
\begin{equation}
 0\le s^{\mathrm{raw}}-s^H
 =\qa[\cb](1-R[\cb,\db])
 \le\qa[\cb]\le1.
 \label{eq:monotone}
\end{equation}
At a fixed threshold, Hard Warrant cannot increase false predictions of
a decision change, but it may increase missed changes. Improvements in
decision accuracy or balanced accuracy are therefore empirical questions.
An unchanged condition predicted by argmax alone does not guarantee
$p^H=\mathbf e_{\db}$: alternative rows may still receive nonzero probability.
Errors in condition estimation and alternative rows remain uncorrected.

The remaining rows depend on the governing rule and non-target condition
states. Under conjunction, a \stateNotSatisfied{} non-target condition
makes every target state yield \decisionNo{}.
If all non-target conditions are \stateSatisfied{}, the three target
states instead yield \decisionYes{}, \decisionNo{}, and
\decisionInsufficient{}, respectively. Thus the original target state and
decision fix one row, while the other rows depend on the remaining conditions.

\section{Data and intervention details}
\label{app:data}

The constructed intervention data contain 6,913 pairs across 1,041 roots,
with 3,680 pairs from ContractNLI and 3,233 from ShARC.
Each root consists of an original case, a target evidence unit, and its
associated condition; each pair records an accepted intervention.
Table~\ref{tab:data} reports the partition of pairs, roots, and connected
components. The same component-disjoint splits are used by all methods.

% CHECK: U71
\begin{table}[H]
\caption{Intervention data splits, with related roots assigned to the same component.}
\label{tab:data}
\centering
\begin{tabular}{lrrrrr}
\toprule
Split & Pairs & Targets & Controls & Roots & Components\\
\midrule
Train & 4,168 & 3,020 & 1,148 & 628 & 116\\
Development & 1,386 & 999 & 387 & 206 & 66\\
Test & 1,359 & 978 & 381 & 207 & 52\\
\midrule
Total & 6,913 & 4,997 & 1,916 & 1,041 & 234\\
\bottomrule
\end{tabular}
\end{table}

% CHECK: END-U71

\subsection{Source adaptations}
\label{app:provenance}

The intervention data use selected cases from structured adaptations of
ContractNLI and ShARC. Each case supplies a governing rule, decision query,
conditions, their states before intervention, an aggregation function, and a
decision. The adaptations also associate evidence units with conditions.

\paragraph{ContractNLI.}
ContractNLI labels hypotheses about a contract and annotates supporting
evidence spans \citep{koreeda2021contractnli}. The structured adaptation
groups hypotheses from the same contract into a case with multiple
conditions. Each hypothesis becomes a condition. The source labels
Entailment, Contradiction, and NotMentioned map to \stateSatisfied{},
\stateNotSatisfied{}, and \stateUnknown{}, respectively. Evidence units
associated with each condition come from the annotated spans. The
adaptation combines the condition states by conjunction to obtain a
decision; this decision is not an original ContractNLI label.

\paragraph{ShARC.}
ShARC provides a rule snippet, question, scenario, and dialogue history
\citep{saeidi2018sharc}. Its structured adaptation represents the rule's
requirements as conditions, associates them with scenario or dialogue
evidence, and specifies how conjunction or disjunction combines them.
The constructor uses these annotated original condition states and evidence
associations; it does not reconstruct the conditions after editing the
evidence. The human semantic audit in Appendix~\ref{app:semantic-audit}
evaluates labels within this inherited representation.

\subsection{Decision labels and complete mappings}
\label{app:decision-labels}

Both sources use the decision space $\Ds$ defined in
Section~\ref{sec:task}. The ContractNLI adaptation maps its labels
\texttt{positive}, \texttt{negative}, and \texttt{insufficient evidence}
to \decisionYes{}, \decisionNo{}, and \decisionInsufficient{}, respectively.
The retained ShARC cases already use these three decisions.

The cases use three-valued conjunction and disjunction.
Under conjunction, any \stateNotSatisfied{} condition yields \decisionNo{};
all \stateSatisfied{} conditions yield \decisionYes{}; otherwise the
decision is \decisionInsufficient{}.
Under disjunction, any \stateSatisfied{} condition yields \decisionYes{};
all \stateNotSatisfied{} conditions yield \decisionNo{}; otherwise the
decision is \decisionInsufficient{}.
After an accepted intervention, the judged target condition state replaces
its original state in $C(x_g)$ while non-target states remain fixed.
The aggregation function then determines the decision after the
intervention and the decision-change label. Applying the same function to
all three possible target states gives $T_g$ in Equation~\ref{eq:target}.
These rule-derived decisions are separate from judging which target state
the edited evidence supports.

\begin{figure}[t]
\centering
\includegraphics[width=\linewidth]{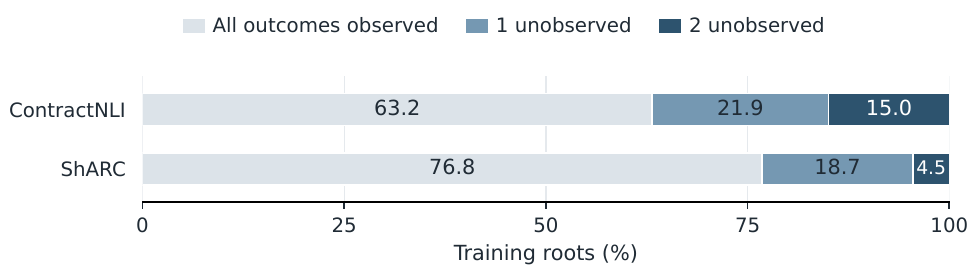}
\caption{\textbf{Additional decisions from complete condition-to-decision mappings.}
Segments show the within-source proportions of training roots with zero,
one, or two decisions absent from the original cases and sampled interventions.}
\label{fig:branch-coverage}
\end{figure}

\label{app:supervision-coverage}
The complete mappings also supervise decisions not observed in the original
cases or sampled interventions. We measure this coverage over all 628
training roots, including invariance controls. For 60.4\% of these roots,
the original case and sampled interventions cover fewer than three target
condition states. For 31.1\%, the complete mapping supplies decisions not
observed in those examples. Figure~\ref{fig:branch-coverage} groups roots
by the number of additional decisions. The original decision counts as
observed even if no sampled intervention produces it. Additional decisions
occur in 36.8\% of ContractNLI roots and 23.2\% of ShARC roots. These
decisions follow from the aggregation function. This comparison measures
supervision coverage; it does not establish a performance gain.

% CHECK: U77

% CHECK: END-U77

\subsection{Intervention construction and corpus composition}
\label{app:interventions}
\label{app:construction-review}

The constructor checks that the target evidence unit is present and that
the original condition states agree with the original decision under the
aggregation function. These checks establish structural consistency, not
independent semantic correctness. The constructor uses source- and
operation-specific prompts for textual rewrites and distractor insertion;
removal deletes the target evidence unit directly. Construction prompts
and configurations will accompany the code release.

Model-assisted judgments assess the target condition state from the
complete evidence after each intervention, both directly and by comparing
the cases before and after the intervention. The judgments do not receive
the expected label or operation name; disagreements receive further review.
Checks confirm evidence associations and preservation of non-target
condition states. Assessments of operation fidelity and preservation of
non-target evidence guide acceptance, review, and sample weighting.
The aggregation function then determines the decision after intervention
and the complete condition-to-decision mapping. Interventions from the
same original case and target evidence unit share this mapping. The rule
and non-target condition states remain fixed across them.

Table~\ref{tab:operation-composition} defines the six target operations and
two controls. For some roots, no valid intervention was obtained for one or
more target operations. Missing interventions are not assigned an unchanged
decision label, while accepted interventions with unchanged decisions still
provide supervision. The table reports retained pairs, not attempted edits
or reasons for missing interventions.

\begin{table}[htbp]
\caption{Intervention operations and retained pair counts.
Target operations may leave the target condition state unchanged.}
\label{tab:operation-composition}
\centering
\begin{tabularx}{.96\textwidth}{lYrrr}
\toprule
Operation & Intended edit & Train & Dev & Test\\
\midrule
Removal & Delete the designated evidence unit. & 616 & 203 & 203\\
Polarity reversal & Reverse the polarity of the same core relation. & 377 & 121 & 127\\
Alternative replacement & Substitute a different fact of the relevant type. & 498 & 165 & 157\\
Irrelevant replacement & Replace the target with content unrelated to the conditions. & 546 & 177 & 177\\
Weakening & Reduce evidential sufficiency without simply negating the assertion. & 519 & 174 & 154\\
Strengthening & Add specificity or support while preserving the intended relation. & 464 & 159 & 160\\
\midrule
Distractor insertion (control) & Insert irrelevant content without replacing the evidence. & 626 & 206 & 205\\
Paraphrase (control) & Change wording while preserving meaning and strength. & 522 & 181 & 176\\
\bottomrule
\end{tabularx}
\end{table}

The six target operations in Table~\ref{tab:operation-composition}
yield 4,997 pairs, and the two controls yield 1,916 pairs. Of the 978
test target pairs, 495 are invariant, 256 are
absorbed, and 227 are propagated. The source breakdown is 506 ContractNLI
and 472 ShARC target pairs. The full test set, including controls, contains
716 ContractNLI and 643 ShARC pairs. Among target interventions that change
the target condition state, 53.0\% leave the decision unchanged. This
fraction describes the constructed intervention pairs; it does not estimate
how often evidence is decision-critical in naturally occurring cases.

Other evidence can preserve the target condition state after its designated
evidence unit is removed. Weakening and strengthening also need not change
the target condition state or decision. The judged state follows the full
evidence after the intervention, and the decision follows the unchanged
aggregation function. No retained root has \stateUnknown{} as its original
target condition state, although \stateUnknown{} appears after some
interventions and is included in complete mappings. Generalization to
original cases with an \stateUnknown{} target condition state therefore
requires further evidence.

Roots linked by source document, rule family, or detected near-duplicates
form connected components assigned wholly to one split, as reported in
Table~\ref{tab:data}. The 349 initial groups contribute 210/69/70 groups
to train/development/test. The final components also link related contexts.
All methods use this partition.

A model-assisted construction review covered 400 pairs: one from each of
the 349 initial groups, plus pairs from under-covered or error-prone
operations. It checked operation fidelity, evidence associations,
preservation of non-target evidence, and target condition states after
intervention, taking the original labels as given. Label errors were
corrected. Invalid evidence associations and clear operation mismatches
were excluded. When an evidence association was invalid, other
interventions from the same original case and target evidence unit were
removed as well. This sample supports error
discovery, not an unbiased estimate of label accuracy.

\subsection{Reference criticality labels}
\label{app:criticality-labels}

Reference criticality labels for roots follow the six target operations in
Table~\ref{tab:operation-composition}. Paraphrases and distractor
insertions are controls and do not contribute to these labels.
An accepted intervention that changes the reference decision establishes
a positive label. Roots with no observed decision change undergo further
review before receiving a negative label. Among the 207 test roots, the 122
with negative labels have 551 accepted target interventions. Of these roots,
36 cover all six operation types, while 86 have incomplete coverage. Even
coverage of all six types does not exhaust the allowed text changes.

For 88 negative roots, the complete mapping gives the original decision
for every target condition state: $T_g(c)=\db_g$ for all $c\in\Cs$.
Under the fixed rule and non-target context, changes to the target
condition therefore cannot alter the decision.
For the remaining 34 roots, reviewers examined the governing rule, the
target and non-target evidence, and the 140 available interventions. They
assessed whether changing only the target evidence could produce a target
condition state that would change the decision required by the rule, while
preserving the rule and non-target condition states. Based on this review,
the 34 roots were labeled non-critical. These judgments reflect an
assessment of the case evidence rather than an exhaustive enumeration of
possible edits. The final evaluation contains 85 critical and 122
non-critical roots. ContractNLI contributes 117 roots, with 41 critical and
76 non-critical labels. ShARC contributes 90 roots, with 44 critical and
46 non-critical labels.

\subsection{Human semantic audit}
\label{app:semantic-audit}

A separate human audit sampled 400 pairs uniformly without replacement
from the 6,913 accepted pairs. Three reviewers independently examined each
pair before discussing disagreements. They saw the governing rule, decision
query, conditions, target evidence unit and its condition, cases before and
after the intervention, and original condition states and decision. The original
labels were assumed correct. The reference condition state and decision
after intervention, complete mapping, model judgments, and other reviewers'
answers were hidden. The intended operation was revealed only after the
reviewers first judged the target condition state.

Reviewers checked the target evidence association, target condition state
after intervention, preservation of non-target condition states, and
operation fidelity. They resolved disagreements against the text and
instructions, recording their reasons. They distinguished unresolved
annotation ambiguity from the task state \stateUnknown{}.
The aggregation function determined decisions after intervention from the
adjudicated target states and supplied non-target states. Pre-correction
labels for the edited cases were revealed only after adjudication.

Table~\ref{tab:semantic-audit} reports validity and agreement with adjudicated
labels before correction, each over 400 pairs. These rates are neither
inter-reviewer agreement nor residual error rates in the experimental data.
Approximate 95\% Wilson intervals for the four rows, in table order, are
$[99.05,100.00]$, $[95.14,98.46]$, $[87.51,93.21]$, and
$[93.91,97.71]$.
The audit is conditional on the supplied original labels and aggregation;
it does not establish correctness of every corpus label.
Identified errors were corrected before all reported experiments.

% CHECK: U65
% Completed human-audit results confirmed by the author on 2026-09-09.
% The author confirmed that the three earlier percentages also match this audit.
% These are pre-correction audit statistics. Experiments use corrected labels.
\begin{table}[htbp]
\caption{Human audit of 400 randomly sampled pairs before correction.
Entries measure validity or agreement with adjudicated labels; original
condition states and decisions are taken as given.}
\label{tab:semantic-audit}
\centering\small
\setlength{\tabcolsep}{9pt}
\begin{tabular}{lrr}
\toprule
Aspect assessed & $n/N$ & Pass / agreement (\%)\\
\midrule
Case structure and evidence association & 400/400 & 100.00\\
Intervention fidelity & 389/400 & 97.25\\
Target condition state & 363/400 & 90.75\\
Decision after intervention & 385/400 & 96.25\\
\bottomrule
\end{tabular}
\end{table}

% CHECK: END-U65

\section{Baseline definitions and adaptations}
\label{app:baselines}

Direct evaluation queries each language model with the same
single-case prompt in Appendix~\ref{app:direct-prompt}. Paired evaluation
instead uses each method's complete system and a supplied intervention pair.
This section specifies the training and paired inputs for each baseline.

Paired evaluation provides the governing rule, decision query, conditions,
aggregation function, target evidence unit and condition, and cases before
and after the intervention. Each method uses the inputs specified below.
The original decision $\db$ is supplied and assumed correct. The target
condition state before the intervention $\cb$ is also available to the
\method{} Verifier, its ablations, and the prompted baseline. Operation
names, reference labels after the intervention, non-target condition
states, and complete mappings are excluded from prediction inputs.
Table~\ref{tab:supervision} summarizes training targets and paired outputs;
Appendix~\ref{app:controls} defines the matched ablations.

% CHECK: U70
\begin{table}[htbp]
\caption{Training supervision and paired outputs.
After-intervention reference labels and complete mappings are not prediction
inputs. Target state denotes target condition state.}
\label{tab:supervision}
\centering\small
\begin{tabularx}{.96\textwidth}{l Y Y}
\toprule
Method & Training supervision & Paired outputs\\
\midrule
Prompted Qwen3.5-4B & None & Target state, decision\\
After-State SFT & Target state, decision & Target state, decision\\
Predicted State Execution & Target-state annotations for its estimator & Target state, decision from aggregation\\
PairCFR & Decisions and pair relation & Decision\\
GenRM & Answer generation, candidate correctness & Decision\\
R$^2$-Guard & SFT estimator, decisions for rule weights & Target state, decision from weighted rules\\
w/o propagation composition & Shared estimator, decisions, complete mappings & Target state, direct decision, conditional decisions\\
w/o branch loss & Shared estimator, observed decisions & Target state, composed decision, conditional decisions\\
w/o change loss & Shared estimator, decisions, complete mappings & Target state, composed decision, conditional decisions\\
w/o Hard Warrant & Shared estimator, decisions, complete mappings & Target state, composed decision, conditional decisions\\
\method{} & Shared estimator, decisions, complete mappings & Target state, composed decision, conditional decisions\\
\bottomrule
\end{tabularx}
\end{table}

% CHECK: END-U70

\subsection{Prompt-only baseline}
\label{app:prompted}

Prompted Qwen3.5-4B receives both cases, the supplied original decision and
target condition state before the intervention, and definitions of the three
condition states and decisions. It reasons about the edit, then predicts the
target condition state and decision after the intervention. No reference
labels after the intervention or worked test examples are supplied.
A fixed serialization of the nine condition-state and decision combinations
gives a normalized distribution from label continuations, rather than
verbalized confidence. Its marginals give $\qa$ and $p$. The same prompt
template and reasoning budget are used throughout; examples remain in the
evaluation when the reasoning budget is exhausted.

\subsection{After-State SFT}

The model reads the complete case after intervention $x^{\mathrm a}$ and
learns to output the target condition state followed by the decision using
next-token cross-entropy. Let $P_\theta$ denote its normalized label-continuation
probabilities. During paired evaluation, decision prediction marginalizes
over the three candidate condition prefixes:
\begin{equation}
 p(d\mid x^{\mathrm a})=
 \sum_{c\in\Cs}P_\theta(c\mid x^{\mathrm a})
                 P_\theta(d\mid x^{\mathrm a},c).
 \label{eq:sft-marginal}
\end{equation}
The prefixes enumerate candidate target condition states, not the reference
state after the intervention. The second factor predicts the next output;
it is not trained against decisions for every candidate target state.

\subsection{GenRM adaptation}
\label{app:genrm}

GenRM jointly trains answer generation and generative verification
\citep{zhang2025genrm}. We use its version without chain-of-thought targets.
For each training pair, the reference decision after the intervention is a
generation target. Each of the three candidate decisions is also a binary
query about its correctness, labeled by $\ind[d=\da]$. The Yes/No
verification response concerns candidate correctness, not the case decision.
The correct candidate receives half of the verification
weight. The two incorrect candidates receive one quarter each.
Response-token losses are averaged within a query, with generation and
class-balanced verification weighted $1/4$ and $3/4$, respectively.
No complete mappings or reference reasoning traces are supplied.

During paired evaluation, GenRM judges each candidate decision using the
case after the intervention. Let $v_\theta(x,d)$ be its Yes probability
normalized over fixed Yes/No continuations for candidate $d$. We normalize
the three candidate scores to obtain the decision distribution:
\begin{equation}
 p^{\mathrm{GenRM}}(d\mid x)=
 \frac{v_\theta(x,d)}{\sum_{d'\in\Ds}v_\theta(x,d')},\qquad
 s^{\mathrm{GenRM}}(x)=1-p^{\mathrm{GenRM}}(\db\mid x).
 \label{eq:genrm-score}
\end{equation}
These normalized scores need not be calibrated. GenRM does not predict
condition states or require a rationale at test time.

\subsection{PairCFR adaptation}
\label{app:paircfr}

PairCFR combines decision classification with supervised contrastive
learning from counterfactual examples \citep{qiu2024paircfr}.
Cases before and after intervention carry their reference decisions. The
original case is deduplicated within each root. Cases with different
decisions before and after intervention appear in the same
contrastive batch. Edits with unchanged decisions remain classification
examples and are not forced into negative pairs.

We retain the authors' contrastive loss\footnote{
\url{https://github.com/Siki-cloud/PairCFR}.}, replacing the encoder
classification representation with the decoder's sequence representation.
The contrastive weight and temperature follow the original NLI setting,
and paired examples share a contrastive batch.
During paired evaluation, the classifier predicts the three-class decision
after the intervention and uses $\db$ to form $s=1-p[\db]$.
It does not output condition states.

\subsection{\texorpdfstring{R$^2$-Guard}{R2-Guard} and predicted-state execution}
\label{app:execution}

R$^2$-Guard combines category likelihoods with weighted logical rules
\citep{kang2025r2guard}. In our adaptation of its probabilistic
formulation\footnote{\url{https://github.com/kangmintong/R-2-Guard}.},
condition states replace safety categories and the known aggregation
function replaces the safety policy. A frozen After-State SFT estimator
supplies condition-state probabilities $q_j(c_j\mid x^{\mathrm a})$ for
conditions $j=1,\ldots,m$ and a directly predicted decision distribution
$p_0(d\mid x^{\mathrm a})$. No annotated non-target condition states are
supplied at prediction.

Let $\boldsymbol c\in\Cs^m$ be a condition-state assignment. For the
three decisions $d_k\in\Ds$, the rule clauses are
$\phi_k(\boldsymbol c,d): A(\boldsymbol c)=d_k\Rightarrow d=d_k$.
With real clause weights $w_k$, the normalized model is
\begin{equation}
 P_w(\boldsymbol c,d\mid x^{\mathrm a})\propto
 p_0(d\mid x^{\mathrm a})
 \prod_{j=1}^m q_j(c_j\mid x^{\mathrm a})
 \exp\!\left(\sum_{k=1}^{3}w_k\ind[\phi_k(\boldsymbol c,d)]\right).
 \label{eq:r2guard}
\end{equation}
Weights are shared by aggregation type and decision, giving six values for
conjunction and disjunction. They are fitted to training decisions with
the neural estimator fixed. Exact summation over $\Cs^m\times\Ds$ gives the
decision marginal. The product of state likelihoods is a modeling
factorization, not a claim of independence between textual conditions.

Predicted State Execution uses its own frozen condition estimator, trained
with the same architecture and training procedure as R$^2$-Guard's
estimator. With $q_j$ denoting its condition-state probabilities, it
directly applies the known aggregation function:
\begin{equation}
 p^{\mathrm{exec}}(d\mid x^{\mathrm a})=
 \sum_{\boldsymbol c\in\Cs^m}
 \left(\prod_{j=1}^m q_j(c_j\mid x^{\mathrm a})\right)
 \ind[A(\boldsymbol c)=d].
 \label{eq:direct-execution}
\end{equation}
It has neither learned rule weights nor a direct-decision prior.
Both methods include \stateUnknown{} and sum over predicted condition
states rather than using annotated states or only the most likely state
for each condition. Each method reports its own estimate of the target
condition state. These comparisons test how systems that combine predicted
states with explicit rule computations perform. The matched ablations in
Appendix~\ref{app:controls} examine propagation composition separately.

\section{Training, selection, and reporting}
\label{app:implementation}
\suppressfloats[t]

\subsection{Two-stage training}
\label{app:objective}

\paragraph{Initialization and conditional queries.}
After-State SFT learns the target condition state and decision from cases
after intervention. The checkpoint with the highest development
decision-change average precision (AP) supplies the frozen estimator
$P_\phi$ and initializes $F_\theta$. The \method{} Verifier and its
ablations share this initialization within each run. The standalone SFT
baseline instead uses development decision negative log-likelihood (NLL)
for selection, as do the second-stage models.

For each pair, three queries ask for the decision under each possible
target condition state while the non-target context stays fixed. The queries
share the same cases before and after the intervention and one forward pass.
Their label logits define $R_i$. Answer positions contain fixed
placeholders; reference answers serve only as loss targets. The frozen
estimator supplies the target condition-state probabilities.

\paragraph{Training objective.}
The second stage supervises each row of $R_i$ with $T_{g(i)}$ and
supervises the composed decision and decision-change predictions with the
observed labels. All accepted interventions pass automated edit checks and
checks that non-target condition states remain unchanged. Their condition
and decision labels are resolved. A pair is marked \texttt{extended} when
at least one assessment of operation fidelity, consistency of non-target
context, or naturalness is flagged or unavailable. Such pairs receive
weight $\omega_i=0.5$ for the decision and decision-change losses; all
other accepted pairs receive $\omega_i=1$. For a microbatch $\mathcal B$,
the losses are
\begin{align}
 \mathcal L_{\mathrm{branch}}&=-\frac{1}{3|\mathcal B|}
   \sum_{i\in\mathcal B}\sum_{c\in\Cs}\log R_i[c,T_{g(i)}(c)],\\
 \mathcal L_{\mathrm{decision}}&=-\frac{\sum_{i\in\mathcal B}\omega_i\log p_i[\da_i]}{
                                          \sum_{i\in\mathcal B}\omega_i},\\
 \mathcal L_{\mathrm{change}}&=-\frac{\sum_{i\in\mathcal B}\omega_i
        [y_i\log s_i+(1-y_i)\log(1-s_i)]}{\sum_{i\in\mathcal B}\omega_i},\\
 \mathcal L_{\mathrm{train}}&=0.5\mathcal L_{\mathrm{branch}}
                   +\mathcal L_{\mathrm{decision}}+0.5\mathcal L_{\mathrm{change}}.
 \label{eq:training-objective}
\end{align}
We assign unit weight to the decision loss and weight 0.5 to each of the
branch and change losses. Matched ablations retain these weights except
when the corresponding loss is removed.
The branch loss weights all three target states and all pairs equally.
Its targets come from the fixed governing rule and non-target context.
The decision and change losses use $p_i$ and $s_i$ after Hard Warrant, as defined in
Equations~\ref{eq:projection} and~\ref{eq:criticality-probability}.
Normalized microbatch losses are averaged over each accumulation window,
excluding padding. Only $\theta$ is updated; $\phi$ remains fixed.
Table~\ref{tab:training} gives the training settings, and
Algorithm~\ref{alg:training} summarizes the two stages.

% CHECK: U53
% Baseline-specific settings remain pending. Matched settings are verified.
\begin{table}[!htb]
\caption{Training and paired evaluation settings for the \method{} Verifier and its ablations.}
\label{tab:training}
\centering\small
\setlength{\tabcolsep}{6pt}
\begin{tabularx}{.94\textwidth}{l Y}
\toprule
Setting & Specification\\
\midrule
\multicolumn{2}{l}{\emph{Verifier training}}\\
Backbone / adaptation & Qwen3.5-4B / LoRA\\
Initialization & Shared After-State SFT checkpoint\\
Condition estimator & Shared and fixed for matched controls\\
LoRA rank / scale / dropout & 32 / 64 / 0.05\\
Trainable modules & Linear-layer adapters, embeddings and output layer\\
Learning rate / schedule & $5\!\times\!10^{-5}$, cosine decay, 3\% warmup\\
Optimizer / weight decay & Adam / 0.1\\
Global / micro batch & 16 / 2\\
Training pair exposure & Two passes\\
Loss weights & $\lambda_{\mathrm{branch}}=0.5$, $\lambda_{\mathrm{change}}=0.5$; decision weight $1$\\
Loss averaging & Equation~\ref{eq:training-objective}\\
Decision composition & Same in training and paired evaluation for each variant\\
Checkpoint count / cadence & 11, every 50 updates and final update\\
\midrule
\multicolumn{2}{l}{\emph{Evaluation and selection}}\\
Input truncation & None\\
Verifier checkpoint selection & Development decision NLL\\
Threshold & One global development threshold after selection\\
\bottomrule
\end{tabularx}
% CHECK: END-U53

\vspace{3pt}
\begin{minipage}{.94\textwidth}
\footnotesize
The condition estimator is learned from the pretrained backbone during
the first training stage and remains fixed during \method{} Verifier training.
\end{minipage}
\end{table}

\begin{algorithm}[!tb]
\caption{Two-stage training of the \method{} Verifier.}
\label{alg:training}
\small
\begin{algorithmic}[1]
\Require Training pairs $z_i$, labels $(\cb_i,\db_i,\ca_i,\da_i,y_i)$,
         mappings $T_{g(i)}$, quality weights $\omega_i$, and development pairs.
\Ensure Parameters $\theta^*$ for direct inference; estimator $\phi$ and
        threshold $\tau^*$ for paired evaluation.
\State Train After-State SFT on cases after intervention
\State Select the SFT checkpoint with highest development decision-change AP
\State Freeze its parameters as $\phi$ and initialize $\theta$ from the same checkpoint
\For{each training batch}
  \State Estimate $\qa_i$ with $P_\phi$
  \State Predict $R_i$ with $F_\theta$ for all target condition states
  \State Apply Hard Warrant and compute $p_i$ and $s_i$
         using Equations~\ref{eq:hard-warrant}--\ref{eq:criticality-probability}
  \State Update $\theta$ with Equation~\ref{eq:training-objective}, keeping $\phi$ fixed
\EndFor
\State Select $\theta^*$ by development decision NLL in Equation~\ref{eq:validation-nll}
\State Select $\tau^*$ by development decision-change BA for the selected model
\State \Return $\theta^*$ for direct inference, and $(\phi,\theta^*,\tau^*)$ for paired evaluation
\end{algorithmic}
\end{algorithm}

\subsection{Checkpoint and threshold selection}
\label{app:selection}

Verifier checkpoints are compared by decision NLL on development target
pairs $\mathcal T_{\mathrm{dev}}$:
\begin{equation}
 \mathcal L_{\mathrm{val}}=-\frac{1}{|\mathcal T_{\mathrm{dev}}|}
 \sum_{i\in\mathcal T_{\mathrm{dev}}}\log p_i[\da_i].
 \label{eq:validation-nll}
\end{equation}
This criterion evaluates each variant's final decision distribution,
including Hard Warrant for variants that use it. The checkpoint with the
lowest NLL supplies $\theta^*$ for direct criticality inference. Exact ties favor
the earlier training step.

For paired evaluation only, a global threshold $\tau^*$ maximizes
development balanced accuracy for $\hat y_i=\ind[s_i\ge\tau^*]$.
Candidate thresholds cover every distinct set of predictions produced by
the development scores, including all-positive and all-negative predictions.
Ties favor the largest candidate threshold. Models without trained checkpoints also receive
a development-selected threshold. Each model and threshold is then fixed
across all paired test metrics, sources, operations, and change regimes.
Direct inference uses generated Yes/No answers and no score threshold.

\subsection{Metrics and reporting}
\label{app:metric-details}
\label{app:reporting}

\paragraph{Direct criticality judgments.}
Accuracy, BA, and Macro-F1 compare the binary prediction $\hat y_g$ with
the reference criticality label $y_g$, with equal root weights.
Critical evidence is the positive class. BA averages recall for the two
classes; Macro-F1 averages their F1 scores. Invalid-answer rate is the
percentage of responses that cannot be parsed as Yes or No.
Invalid answers remain in the denominator: each counts as an accuracy
error and a missed instance of its reference class for recall and F1.
They are not mapped to No. Predicted State Execution sometimes reports
contradictions or says it cannot judge the case. R$^2$-Guard sometimes
analyzes the evidence without providing the requested Yes/No answer.
Source results compute these metrics within each source before averaging
across runs.

\paragraph{Paired prediction.}
Decision-change BA compares $\ind[s_i\ge\tau^*]$ with
$y_i=\ind[\da_i\ne\db_i]$. Predictions of the target condition state and
decision after intervention are the argmax of $\qa_i$ and $p_i$,
respectively. Condition-state metrics assess only the target condition.
Macro-F1 for each prediction averages over its three classes.
Absorbed and propagated accuracies restrict decision accuracy to the
corresponding reference subsets. Control FPR is the proportion of control
pairs incorrectly predicted to change the decision. AP used for first-stage
selection is non-interpolated Average Precision, with tied scores grouped.

\paragraph{Conditional decision predictions.}
Each row of $R_i$ is evaluated against $T_{g(i)}(c)$ before Hard Warrant.
\emph{Decision Acc.\ (Alternative States)} averages correctness over the
two states $c\ne\cb_i$; \emph{All Three Decisions Correct} requires correct
argmax predictions for all three states. Both metrics first average over
target pairs within a root, then equally across roots.

\paragraph{Aggregation across runs.}
Each run uses the same configuration. Metrics are computed separately
for each run before averaging, not from ensembled predictions.
All percentages and their standard deviations use the same scale.
Argmax ties follow the label orders in Section~\ref{sec:task}; an F1 score
with zero denominator is set to zero. A subgroup metric lacking a required
reference class is undefined, not zero.

\subsection{Direct criticality prompt}
\label{app:direct-prompt}

All models receive the same single-case prompt. The input fields are
shown with placeholders below, followed by the exact evaluation question.
No condition-state labels, decision labels, intervened evidence, or
intervention outcomes are supplied.

\begin{quote}
\noindent
\textbf{Original case:} \texttt{<original case>}\\
\textbf{Decision rule:} \texttt{<decision rule>}\\
\textbf{All conditions:} \texttt{<condition descriptions>}\\
\textbf{Target condition:} \texttt{<target condition>}\\
\textbf{Target evidence location:} \texttt{<target evidence location>}\\
\textbf{Aggregation rule:} \texttt{<aggregation rule>}

\noindent
Would changing only the target evidence, while keeping all other case
facts and decision rules fixed, be capable of changing the final decision?\\
Answer exactly Yes or No.
\end{quote}

The question asks whether a change to the target evidence could alter
the decision required by the rule. Reference labels follow the intervention
and review protocol in Appendix~\ref{app:criticality-labels}.

\section{Controlled comparisons and error analysis}
\label{app:controls}

Within each run, the ablations share the frozen condition estimator,
SFT initialization, paired inputs, and training exposure. Their paired
condition predictions are therefore identical. Table~\ref{tab:controls}
evaluates downstream paired decisions, while
Table~\ref{tab:direct-ablations} tests the trained language models on
direct criticality judgments.

\subsection{Matched ablations}

\paragraph{Composition ablation.}
\emph{w/o propagation composition} retains the three conditional decision
queries and branch loss but does not use their predictions to compute the
decision after intervention. Instead, a direct decision query produces a
three-class distribution $u_i$. To preserve the probability assigned to
the original decision by Hard Warrant's fixed row, this variant uses
\begin{equation}
 p_i^{\mathrm{flat}}=\qa_i[\cb_i]\mathbf e_{\db_i}
                  +(1-\qa_i[\cb_i])u_i,\qquad
 s_i^{\mathrm{flat}}=1-p_i^{\mathrm{flat}}[\db_i].
 \label{eq:flat-ablation}
\end{equation}
Here $\mathbf e_{\db_i}$ is the one-hot distribution for the original
decision. The variant retains the conditional decision queries and loss
coefficients, but the decision and change losses use $p_i^{\mathrm{flat}}$.
No row of $R_i$ enters this distribution. The comparison therefore tests
propagation composition while retaining branch supervision and the known
relation between the original target state and decision.

\paragraph{Branch-loss ablation.}
\emph{w/o branch loss} retains the three conditional decision
queries, the fixed condition estimator, and Hard Warrant. It changes only
the branch-loss coefficient in Equation~\ref{eq:training-objective} to
zero. The decision and change losses still train the conditional decision
predictions through propagation composition. No complete-mapping label
contributes directly to the loss.

\paragraph{Change-loss ablation.}
\emph{w/o change loss} sets only the coefficient of
$\mathcal L_{\mathrm{change}}$ to zero. Branch supervision, decision
supervision, propagation composition, and Hard Warrant are retained.
This comparison isolates the additional binary change objective from
supervision of the full decision distribution.

\paragraph{Hard Warrant ablation.}
\emph{w/o Hard Warrant} uses $p_i=\qa_i R_i$ throughout training
and paired prediction, with all three training losses retained.
Each ablation starts from the same run's SFT initialization and uses
development decision NLL for checkpoint selection.

\subsection{Predicted probabilities of target condition and decision changes}
\label{app:score-distributions}

\begin{figure}[H]
\centering
\includegraphics[width=0.8\linewidth]{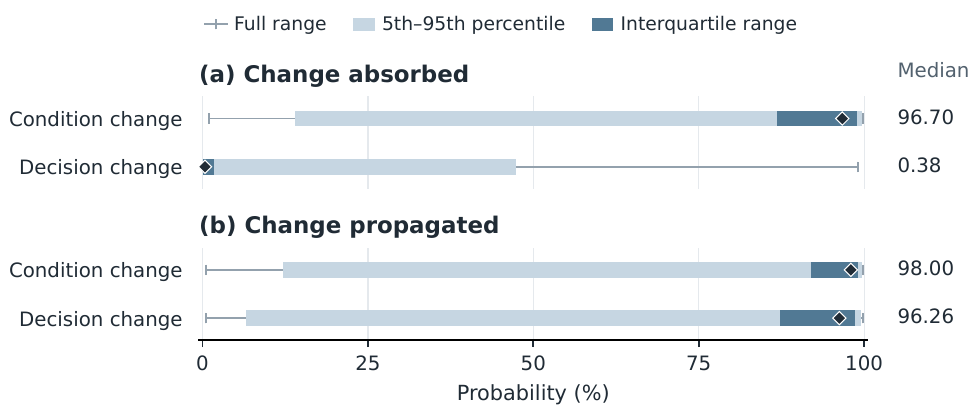}
\caption{\textbf{Predicted probabilities of target condition and decision changes.}
Shown for test interventions whose reference target condition state changes.
Lines show full ranges; light bands show the 5th--95th percentiles;
dark bands show interquartile ranges; markers show medians. Runs, roots,
and interventions within roots receive equal weight in each group.
Bands are not confidence intervals.}
\label{fig:raw-score-density}
\end{figure}

Figure~\ref{fig:raw-score-density} uses all test target interventions
whose reference target condition state changes, across the three main-model
runs. Prediction errors remain included. It compares
$1-\qa_i[\cb_i]$, the predicted probability of a target condition change,
with the decision-change score $s_i$. The absorbed and propagated groups
are normalized separately. Within each group, runs have equal weight, as
do roots within a run and interventions within a root. Both groups have
high median probabilities of a target condition change, but their median
decision-change scores differ. These distributions describe predictions;
the ablations test the contribution of propagation composition.
The bound $s_i\leq1-\qa_i[\cb_i]$ follows from Hard Warrant, independently
of the observed separation.

\Needspace{19\baselineskip}
\subsection{Hard Warrant at a fixed checkpoint}
\label{app:errors}
\label{sec:correction-results}

\begingroup
\setlength{\intextsep}{5pt}
\setlength{\columnsep}{10pt}
\setlength{\abovecaptionskip}{5pt}
\begin{wrapfigure}[17]{r}{0.43\textwidth}
\centering
\includegraphics[width=\linewidth]{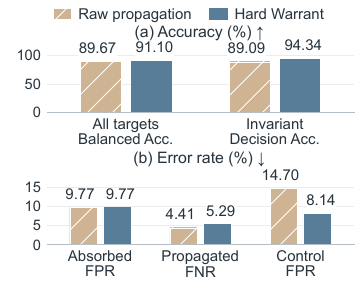}
\caption{Hard Warrant with checkpoint and threshold fixed.
Point estimates.}
\label{fig:factual-correction}
\end{wrapfigure}

The diagnostic compares $p_i^{\mathrm{raw}}=\qa_i R_i$ and
$p_i^H=\qa_i R_i^H$ at the same selected checkpoint and threshold,
without retraining or threshold reselection.
Figure~\ref{fig:factual-correction} reports point estimates: Hard Warrant
raises decision-change BA from 89.67\% to 91.10\% and invariant decision
accuracy from 89.09\% to 94.34\%. Control FPR falls from 14.70\% to 8.14\%,
but the missed-change rate on propagated targets rises from 4.41\% to 5.29\%.
This trade-off is consistent with the score reduction in Appendix~\ref{app:proof}.

This comparison isolates the effect of the constraint on paired predictions.
It differs from the ablation above, which trains and selects a separate
model without Hard Warrant. That ablation tests the constraint's contribution
during training to direct criticality judgments.
Direct inference does not execute the constraint.

\WFclear
\endgroup

\section{Results by source, operation, and change regime}
\label{app:subgroup-results}
\label{app:breakdowns}

The first comparison reports direct criticality judgments by source.
The remaining comparisons evaluate supplied intervention pairs and reuse
the models and global thresholds selected for paired evaluation.

\subsection{Direct criticality judgments by source}
\label{app:direct-by-source}

Figure~\ref{fig:direct-criticality-by-source} shows direct criticality
balanced accuracy by source. Table~\ref{tab:direct-criticality-by-source}
expands the comparison in Section~\ref{sec:main-results} and the ablations
in Section~\ref{sec:controls} to all three metrics. Each source uses the
same runs as the overall evaluation, with metrics computed within each
source before averaging across runs.

\begingroup
\setlength{\intextsep}{5pt}
\begin{figure}[H]
\centering
\includegraphics[width=0.60\textwidth]{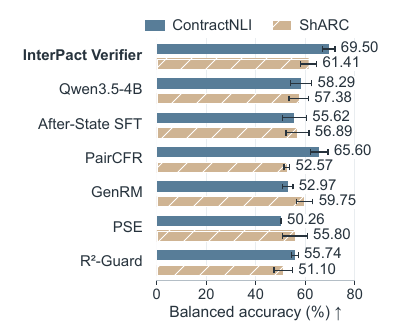}
\caption{Direct criticality balanced accuracy by source. Error bars show
standard deviations across three runs.}
\label{fig:direct-criticality-by-source}
\end{figure}

% Generated from tables/direct_criticality_by_source_results.json.
% Every model includes three runs; GenRM/PSE/R2-Guard reports are reconciled.
\begin{table}[H]
\caption{Direct criticality judgments and training ablations by source
(\%).}
\label{tab:direct-criticality-by-source}
\centering\small
\alignresultcells{00.00}
\setlength{\tabcolsep}{6pt}
\begin{tabular}{lccc}
\toprule
Method & Acc. $\uparrow$ & BA $\uparrow$ & Macro-F1 $\uparrow$\\
\midrule
\multicolumn{4}{c}{\textit{ContractNLI: 117 roots}}\\
\midrule
Qwen3.5-4B & \runresult{62.11}{3.56} & \runresult{58.29}{4.35} & \runresult{58.19}{4.13}\\
After-State SFT & \runresult{66.67}{1.48} & \runresult{55.62}{4.88} & \runresult{51.70}{10.68}\\
PairCFR & \runresult{65.53}{4.86} & \runresult{65.60}{3.62} & \runresult{64.15}{4.25}\\
GenRM & \runresult{66.38}{0.99} & \runresult{52.97}{2.06} & \runresult{46.57}{4.42}\\
Predicted State Execution & \runresult{47.29}{0.49} & \runresult{50.26}{0.06} & \runresult{47.70}{0.07}\\
R$^2$-Guard & \runresult{54.42}{0.99} & \runresult{55.74}{1.41} & \runresult{54.18}{0.71}\\
\method{} & \runresult{73.50}{2.57} & \runresult{69.50}{2.54} & \runresult{70.01}{2.71}\\
\midrule
w/o propagation composition & \runresult{71.22}{0.98} & \runresult{61.00}{0.76} & \runresult{60.32}{0.79}\\
w/o branch loss & \runresult{75.78}{0.99} & \runresult{70.50}{1.41} & \runresult{71.49}{1.41}\\
w/o Hard Warrant & \runresult{64.39}{0.50} & \runresult{50.49}{1.03} & \runresult{42.62}{2.42}\\
w/o change loss & \runresult{67.52}{2.26} & \runresult{55.53}{2.06} & \runresult{52.16}{2.33}\\
\midrule
\multicolumn{4}{c}{\textit{ShARC: 90 roots}}\\
\midrule
Qwen3.5-4B & \runresult{57.41}{3.90} & \runresult{57.38}{3.94} & \runresult{57.37}{3.95}\\
After-State SFT & \runresult{57.41}{4.49} & \runresult{56.89}{4.64} & \runresult{54.23}{6.18}\\
PairCFR & \runresult{52.22}{1.11} & \runresult{52.57}{1.17} & \runresult{51.18}{0.88}\\
GenRM & \runresult{60.37}{3.21} & \runresult{59.75}{3.31} & \runresult{56.50}{4.91}\\
Predicted State Execution & \runresult{55.93}{5.13} & \runresult{55.80}{5.02} & \runresult{55.87}{5.09}\\
R$^2$-Guard & \runresult{51.11}{3.85} & \runresult{51.10}{3.74} & \runresult{51.01}{3.86}\\
\method{} & \runresult{61.48}{3.20} & \runresult{61.41}{3.21} & \runresult{61.40}{3.21}\\
\midrule
w/o propagation composition & \runresult{50.74}{2.79} & \runresult{50.20}{2.84} & \runresult{47.26}{3.54}\\
w/o branch loss & \runresult{48.15}{0.64} & \runresult{47.79}{0.63} & \runresult{46.60}{0.54}\\
w/o Hard Warrant & \runresult{51.48}{1.28} & \runresult{50.77}{1.22} & \runresult{45.54}{0.43}\\
w/o change loss & \runresult{47.78}{3.85} & \runresult{47.58}{3.89} & \runresult{47.22}{4.11}\\
\bottomrule
\end{tabular}
\end{table}

\endgroup

\Needspace{7\baselineskip}
\subsection{Paired results by intervention type}
\label{sec:intervention-results}

In Table~\ref{tab:operation-results}, \method{} has the highest mean
decision-change balanced accuracy for all six operations and decision
accuracy for five. On weakening, GenRM has higher decision accuracy,
92.64\% versus 91.77\%.

% CHECK: U67
\begin{table}[!htbp]
\caption{Decision-change and decision accuracy by target operation (\%).}
\label{tab:operation-results}
\centering\small
\alignresultcells{0.00}
\setlength{\tabcolsep}{2pt}
\begin{tabular*}{\textwidth}{@{\hspace{3pt}\extracolsep{\fill}}lccccc>{\alignresultcells{00.00}}c@{\hspace{3pt}}}
\toprule
Method & Removal & \shortstack{Polarity\\reversal} & \shortstack{Alternative\\replacement} & \shortstack{Irrelevant\\replacement} & Weakening & Strengthening\\
\midrule
\rowcolor{tablegroupblue}[0pt][0pt]
\multicolumn{7}{l}{\emph{Decision Change Balanced Acc. $\uparrow$}}\\
Prompted Qwen3.5-4B & \runresult{78.85}{1.25} & \runresult{90.78}{0.76} & \runresult{78.22}{2.50} & \runresult{76.33}{3.46} & \runresult{76.66}{1.11} & \runresult{78.90}{0.19}\\
After-State SFT & \runresult{86.44}{1.00} & \runresult{92.22}{0.66} & \runresult{\secondresult{83.90}}{3.35} & \runresult{88.97}{2.03} & \runresult{77.99}{1.34} & \runresult{78.97}{7.79}\\
Predicted State Execution & \runresult{85.86}{1.70} & \runresult{90.89}{1.91} & \runresult{80.67}{1.66} & \runresult{89.24}{1.20} & \runresult{78.99}{1.11} & \runresult{\secondresult{86.44}}{5.16}\\
% PairCFR SDs provided by the author on 2026-09-26.
PairCFR & \runresult{84.48}{3.52} & \runresult{\secondresult{92.24}}{0.82} & \runresult{80.45}{1.47} & \runresult{85.55}{0.98} & \runresult{71.77}{2.94} & \runresult{64.29}{8.33}\\
GenRM & \runresult{85.11}{0.53} & \runresult{89.80}{1.16} & \runresult{81.36}{3.02} & \runresult{86.24}{1.31} & \runresult{\secondresult{84.71}}{4.18} & \runresult{81.39}{8.50}\\
R$^2$-Guard & \runresult{\secondresult{87.13}}{0.36} & \runresult{91.90}{1.72} & \runresult{81.47}{2.51} & \runresult{\secondresult{89.64}}{0.82} & \runresult{76.75}{5.73} & \runresult{75.36}{10.92}\\
\midrule
\method{} & \runresult{\bestresult{90.06}}{0.43} & \runresult{\bestresult{93.35}}{2.22} & \runresult{\bestresult{86.24}}{3.28} & \runresult{\bestresult{90.69}}{0.52} & \runresult{\bestresult{86.68}}{3.42} & \runresult{\bestresult{90.87}}{5.38}\\
\midrule
\rowcolor{tablegroupblue}[0pt][0pt]
\multicolumn{7}{l}{\emph{Decision Accuracy $\uparrow$}}\\
Prompted Qwen3.5-4B & \runresult{67.00}{2.26} & \runresult{83.73}{2.77} & \runresult{69.43}{1.69} & \runresult{65.91}{0.33} & \runresult{72.94}{0.37} & \runresult{85.42}{0.36}\\
After-State SFT & \runresult{86.37}{0.57} & \runresult{\secondresult{87.93}}{0.45} & \runresult{\secondresult{83.65}}{0.97} & \runresult{86.82}{0.86} & \runresult{87.45}{0.37} & \runresult{89.79}{0.36}\\
Predicted State Execution & \runresult{84.89}{1.99} & \runresult{86.88}{0.91} & \runresult{82.59}{0.97} & \runresult{86.63}{1.18} & \runresult{85.06}{1.30} & \runresult{87.50}{1.88}\\
PairCFR & \runresult{80.30}{0.99} & \runresult{87.40}{0.79} & \runresult{78.34}{1.10} & \runresult{82.49}{1.13} & \runresult{81.82}{0.65} & \runresult{83.75}{1.08}\\
GenRM & \runresult{85.55}{1.42} & \runresult{86.61}{0.79} & \runresult{83.01}{0.97} & \runresult{86.82}{1.30} & \runresult{\bestresult{92.64}}{0.37} & \runresult{\secondresult{96.67}}{0.36}\\
R$^2$-Guard & \runresult{\secondresult{87.52}}{1.42} & \runresult{87.14}{1.20} & \runresult{\secondresult{83.65}}{2.05} & \runresult{\secondresult{88.51}}{1.42} & \runresult{88.96}{0.65} & \runresult{89.38}{1.25}\\
\midrule
\method{} & \runresult{\bestresult{90.80}}{1.42} & \runresult{\bestresult{88.71}}{1.20} & \runresult{\bestresult{87.26}}{1.27} & \runresult{\bestresult{90.96}}{2.04} & \runresult{\secondresult{91.77}}{0.75} & \runresult{\bestresult{98.54}}{0.36}\\
\bottomrule
\end{tabular*}
\end{table}

% CHECK: END-U67

\subsection{Paired results by source and controls}
\label{app:paired-source-controls}

Table~\ref{tab:source-controls} compares \method{} with GenRM by source
and on control interventions, using the same selected models and thresholds.
Control FPR measures false decision changes on paraphrases and distractor
insertions.

\begin{table}[htbp]
\caption{Decision-change and decision accuracy by source, with control
false-positive rates (\%).}
\label{tab:source-controls}
\centering\small
\alignresultcells{0.00}
\setlength{\tabcolsep}{2.5pt}
\scalebox{0.8}{
\begin{tabular*}{.98\textwidth}{@{\hspace{3pt}\extracolsep{\fill}}lcccc>{\alignresultcells[0.00]{0.00}}c@{\hspace{3pt}}}
\toprule
 & \multicolumn{2}{c}{ContractNLI $\uparrow$}
 & \multicolumn{2}{c}{ShARC $\uparrow$}
 & Controls $\downarrow$\\
\cmidrule(lr){2-3}\cmidrule(lr){4-5}
Method & \shortstack{Decision Change\\Balanced Acc.} & Decision Acc.
 & \shortstack{Decision Change\\Balanced Acc.} & Decision Acc. & FPR\\
\midrule
GenRM & \runresult{85.71}{1.42} & \runresult{87.48}{1.01}
 & \runresult{89.41}{0.62} & \runresult{89.48}{0.49} & \runresult{0.09}{0.15}\\
\method{} & \runresult{92.25}{0.62} & \runresult{93.08}{1.03}
 & \runresult{89.31}{0.92} & \runresult{89.62}{0.85} & \runresult{6.39}{1.60}\\
\bottomrule
\end{tabular*}
}
\end{table}

Table~\ref{tab:source-results} extends the source comparison to all methods.

% CHECK: U68
\begin{table}[!htb]
\caption{Paired results by source (\%).
BA: decision-change balanced accuracy; D-Acc.: decision accuracy after
intervention; C-F1: target condition-state Macro-F1.
N/A: unavailable output.}
\label{tab:source-results}
\centering\small
\setlength{\tabcolsep}{2.5pt}
\scalebox{0.85}{
\begin{tabular*}{\textwidth}{@{\hspace{3pt}\extracolsep{\fill}}lrrrrrr@{\hspace{3pt}}}
\toprule
 & \multicolumn{3}{c}{ContractNLI $\uparrow$} & \multicolumn{3}{c}{ShARC $\uparrow$}\\
\cmidrule(lr){2-4}\cmidrule(lr){5-7}
Method & BA & D-Acc. & C-F1 & BA & D-Acc. & C-F1\\
\midrule
Prompted Qwen3.5-4B & \runresult{80.70}{2.36} & \runresult{68.77}{1.03} & \runresult{58.35}{0.84} & \runresult{84.46}{0.49} & \runresult{78.18}{0.42} & \runresult{70.80}{0.63}\\
After-State SFT & \runresult{87.39}{0.35} & \runresult{87.48}{1.35} & \runresult{84.86}{1.08} & \runresult{86.96}{0.67} & \runresult{86.37}{1.50} & \runresult{81.78}{1.54}\\
Predicted State Execution & \runresult{86.02}{1.95} & \runresult{86.10}{1.09} & \runresult{85.12}{0.54} & \runresult{86.25}{0.72} & \runresult{84.96}{1.60} & \runresult{81.66}{1.80}\\
% PairCFR SDs provided by the author on 2026-09-26.
PairCFR & \runresult{85.43}{0.53} & \runresult{83.40}{0.59} & N/A & \runresult{81.92}{0.73} & \runresult{80.72}{0.37} & N/A\\
GenRM & \runresult{85.71}{1.42} & \runresult{87.48}{1.01} & N/A & \runresult{\bestresult{89.41}}{0.62} & \runresult{\secondresult{89.48}}{0.49} & N/A\\
R$^2$-Guard & \runresult{\secondresult{88.36}}{3.11} & \runresult{\secondresult{89.13}}{1.10} & \runresult{\secondresult{85.54}}{2.67} & \runresult{85.97}{1.13} & \runresult{85.88}{0.86} & \runresult{\bestresult{82.47}}{2.52}\\
\method{} & \runresult{\bestresult{92.25}}{0.62} & \runresult{\bestresult{93.08}}{1.03} & \runresult{\bestresult{86.78}}{1.73} & \runresult{\secondresult{89.31}}{0.92} & \runresult{\bestresult{89.62}}{0.85} & \runresult{\secondresult{82.02}}{1.29}\\
\bottomrule
\end{tabular*}
}
\end{table}

% CHECK: END-U68

The paired decision-accuracy advantage over GenRM is concentrated in
ContractNLI: +5.60 percentage points, compared with +0.14 on ShARC.
On ShARC, GenRM has slightly higher mean decision-change balanced accuracy,
at 89.41\% versus 89.31\%.
On ContractNLI, recall for \decisionInsufficient{} is 85.78\% for
\method{} versus 63.11\% for GenRM, with fewer cases mistaken for
\decisionNo{}.

Across all 978 target pairs, \method{} and GenRM have decision-change
recalls of 94.57\% and 88.25\%, respectively. Their false-positive rates
on unchanged targets are similar, at 12.92\% and 12.96\%.
On control interventions, GenRM has a lower FPR: 0.09\%, compared with
6.39\% for \method{}. Higher overall accuracy therefore does not imply
fewer errors in every setting.

Table~\ref{tab:regime-errors} separates incorrect decision-change
predictions when the target condition state is unchanged, incorrect changes
after an absorbed condition change, and missed decision changes required by
the rule. Unlike Control FPR, errors on absorbed targets test whether a
real target condition change is mistaken for a decision-critical one. These
rates complement the decision accuracies by change regime in
Table~\ref{tab:controls}.

% CHECK: U58
\begin{table}[htbp]
\caption{Decision-change errors by reference-defined regime (\%).
Each model uses its global development-selected threshold.}
\label{tab:regime-errors}
\centering\small
\scalebox{0.9}{
\begin{tabular}{lrrrr}
\toprule
Method & Invariant & Absorbed & Propagated & Controls\\
 & FPR $\downarrow$ & FPR $\downarrow$ & FNR $\downarrow$ & FPR $\downarrow$\\
\midrule
Prompted Qwen3.5-4B & \runresult{22.29}{2.47} & \runresult{20.70}{3.10} & \runresult{13.07}{2.69} & \runresult{\secondresult{5.51}}{1.98}\\
After-State SFT & \runresult{18.72}{4.45} & \runresult{13.80}{3.83} & \runresult{8.52}{3.31} & \runresult{15.22}{2.33}\\
Predicted State Execution & \runresult{21.62}{3.43} & \runresult{12.37}{2.83} & \runresult{9.10}{1.42} & \runresult{19.07}{3.93}\\
% PairCFR SDs provided by the author on 2026-09-26.
PairCFR & \runresult{24.04}{0.61} & \runresult{15.23}{0.39} & \runresult{11.45}{0.76} & \runresult{16.80}{0.26}\\
GenRM & \runresult{\bestresult{11.58}}{1.63} & \runresult{15.62}{2.38} & \runresult{11.75}{1.67} & \runresult{\bestresult{0.09}}{0.15}\\
R$^2$-Guard & \runresult{\secondresult{13.80}}{3.92} & \runresult{12.37}{2.15} & \runresult{12.19}{2.87} & \runresult{11.29}{2.98}\\
\midrule
w/o propagation composition & \runresult{15.42}{5.38} & \runresult{\bestresult{10.81}}{2.35} & \runresult{9.10}{0.67} & \runresult{6.82}{2.78}\\
w/o branch loss & \runresult{14.68}{2.53} & \runresult{13.93}{0.81} & \runresult{7.20}{1.99} & \runresult{5.60}{1.45}\\
w/o change loss & \runresult{16.16}{3.93} & \runresult{14.58}{5.97} & \runresult{\secondresult{6.61}}{2.64} & \runresult{7.61}{1.39}\\
w/o Hard Warrant & \runresult{19.12}{1.53} & \runresult{15.36}{2.39} & \runresult{6.75}{2.26} & \runresult{13.74}{3.04}\\
\method{} & \runresult{13.87}{0.51} & \runresult{\secondresult{11.07}}{1.37} & \runresult{\bestresult{5.43}}{1.99} & \runresult{6.39}{1.60}\\
\bottomrule
\end{tabular}
}
\end{table}

% CHECK: END-U58

\FloatBarrier

\section{Extended related work}
\label{app:related}

% CHECK: U2
Belief-R evaluates conclusion revision after additional contextual premises,
including the tension between updating and retaining prior inferences
\citep{wilie2024belief}. DeltaLogic constructs minimal premise-edit episodes
and measures revised-label accuracy, inertia, over-flips, and abstention
\citep{dhanda2026deltalogic}. The Evidence Intervention Constructor records
the target condition state and decision before and after each accepted
intervention. These records distinguish interventions that leave the target
condition unchanged, change it without changing the decision, or change
both. The complete condition-to-decision mapping also gives the decision
for each possible target condition state under the fixed governing rule and
non-target condition states. We use this mapping to train conditional
decision predictions and evaluate them separately from the decision after
intervention.
% CHECK: END-U2

% CHECK: U3
Process supervision \citep{lightman2024verify} and ProcessBench
\citep{zheng2025processbench} focus on the correctness of intermediate
reasoning. Premise-augmented chains expose dependencies for error
identification \citep{mukherjee2025parc}. Our task identifies evidence whose
allowed changes can alter the decision under the governing rule.
In paired evaluation, the GenRM and R$^2$-Guard adaptations in
Appendix~\ref{app:baselines} compare candidate-decision verification and
explicit rule execution with propagation composition.
% CHECK: END-U3

% CHECK: U4
ContextCite attributes a fixed response to the context used to generate it
\citep{cohenwang2024contextcite}. Our criticality labels concern whether an
allowed evidence change can alter the decision required by the supplied rule.
Controlled variations expose reasoning behavior that aggregate accuracy can
miss \citep{mirzadeh2025symbolic}, while paired counterfactual examples support
contrastive learning \citep{qiu2024paircfr}. \citet{huyuk2025feedback}
use paired factual/counterfactual questions and consistency-based feedback
for fine-tuning, with an objective that goes beyond individual-answer
accuracy. We additionally supervise conditional decision predictions with
complete condition-to-decision mappings and combine them with target
condition-state probabilities estimated from the case after intervention.
Hard Warrant uses the supplied target condition state and decision before
intervention during paired training and evaluation.
% CHECK: END-U4

% CHECK: U5
Concept bottleneck models predict through intermediate concepts and
explicitly distinguish this computation from auxiliary concept prediction
\citep{koh2020cbm}. Work on concept interventions
\citep{shin2023intervention}, probabilistic concepts \citep{kim2023probcbm},
and mechanistically derived concepts \citep{desantis2026mechanistic}
examines different aspects of this interface. Counterfactual concept
bottlenecks also support concept interventions and generate concept changes
for alternative predictions \citep{dominici2025counterfactual}.
The Evidence Intervention Constructor derives a complete
condition-to-decision mapping from the governing rule and non-target
condition states. The \method{} Verifier learns conditional decision
predictions from the intervention pairs. We evaluate these predictions
separately from the estimated target condition state and decision after
intervention. Direct criticality inference uses the trained language
model alone, without supplied states or decisions. Our matched ablations
test propagation composition while holding the condition estimator and
branch loss fixed. They also test the contribution of branch loss to
direct criticality judgments from a single case.

% CHECK: END-U5

\end{document}